\documentclass[journal]{IEEEtran}
\usepackage{caption}
\usepackage{soul,framed} %,caption
\usepackage[table]{xcolor}
\definecolor{navy}{RGB}{0, 0, 128}
\definecolor{steelblue}{RGB}{30, 144, 255}

\colorlet{shadecolor}{yellow}
\usepackage[pdftex]{graphicx}
\graphicspath{{../pdf/}{../jpeg/}}
\DeclareGraphicsExtensions{.pdf,.jpeg,.png}

\usepackage[cmex10]{amsmath}

\usepackage{array}
\usepackage{mdwmath}
\usepackage{mdwtab}
\usepackage{eqparbox}
\usepackage{url}

\usepackage{graphicx}
\usepackage{subcaption}
\usepackage{tabularx} 
\usepackage{graphicx}
\usepackage{amsmath}
\usepackage{array}
\usepackage{booktabs}
\usepackage{color}
\usepackage{multirow}
\usepackage{colortbl}

\usepackage[numbers,sort&compress]{natbib} 
\usepackage{pifont}  % For special symbols like the black solid circle
\usepackage{wasysym} 

\usepackage{mdwlist} % For properly handling lists
 
\usepackage{algorithmic}

\usepackage[ruled,linesnumbered]{algorithm2e}
\usepackage{enumerate}
\usepackage{orcidlink}
\usepackage{hyperref}

\usepackage{graphicx} 
\usepackage[export]{adjustbox} 

\def\BibTeX{{\rm B\kern-.05em{\sc i\kern-.025em b}\kern-.08em
    T\kern-.1667em\lower.7ex\hbox{E}\kern-.125emX}}

\usepackage{xpatch}
\xpatchcmd{\thebibliography}{\list}{\small\list}{}{}
\usepackage{amssymb} 

\usepackage{pifont}

\begin{document}

\bstctlcite{IEEEexample:BSTcontrol}
    \title{Enhancing Trust Management System for Connected Autonomous Vehicles Using Machine Learning Methods: A Survey}
  \author{Qian~Xu,~\orcidlink{0000-0002-3374-4108}
      Lei~Zhang(Member) ~\orcidlink{0000-0003-1961-3176}
      Yixiao~Liu,~\orcidlink{0000-0003-1961-3176}

  \thanks{This work was supported by the “Shanghai Collaborative Innovation Research Center for Multi-network and multi-modal Rail Transit, the Fundamental Research Funds for the Central Universities(2023-4-ZD-03) and the National Natural Science Foundation of China (No.52432012). (Corresponding authors: Qian Xu and Lei Zhang).}
  \thanks{Qian Xu and Lei Zhang are with the Key Laboratory of Road and Traffic Engineering, Ministry of Education, Tongji University, Shanghai, China; College of Transportation, Tongji University, China. (e-mail: xuqian0403@tongji.edu.cn;reizhg@tongji.edu.cn).}% <-this % stops a space
  \thanks{Lei Zhang and Yixiao Liu are with Shanghai Research Institute for Intelligent Autonomous Systems, Tongji University, Shanghai, China.}}

%\markboth{IEEE TRANSACTIONS ON Intelligent Transportation Systems, VOL.~X, NO.~X, ~X}{Roberg \MakeLowercase{\textit{et al.}}: High-Efficiency Diode and Transistor Rectifiers}

%The paper headers
\markboth{IEEE TRANSACTIONS ON XXX, VOL.~X, NO.~X, ~X}{Roberg \MakeLowercase{\textit{et al.}}: High-Efficiency Diode and Transistor Rectifiers}

% ====================================================================
\maketitle

% === ABSTRACT ====================================================================
% =================================================================================
\begin{abstract}
%\boldmath
Connected Autonomous Vehicles (CAVs) operate in dynamic, open, and multi-domain networks, rendering them vulnerable to various threats. Trust Management Systems (TMS) systematically organize essential steps in the trust mechanism, identifying malicious nodes against internal threats and external threats, as well as ensuring reliable decision-making for more cooperative tasks. Recent advances in machine learning (ML) offer significant potential to enhance TMS, especially for the strict requirements of CAVs, such as CAV nodes moving at varying speeds, and opportunistic and intermittent network behavior. Those features distinguish ML-based TMS from social networks, static IoT, and Social IoT.
This survey proposes a novel three-layer ML-based TMS framework for CAVs in the vehicle-road-cloud integration system, i.e., trust data layer, trust calculation layer and trust incentive layer. A six-dimensional taxonomy of objectives is proposed. Furthermore, the principles of ML methods for each module in each layer are analyzed. Then, recent studies are categorized based on traffic scenarios that are against the proposed objectives. Finally, future directions are suggested, addressing the open issues and meeting the research trend. We maintain an active repository that contains up-to-date literature and
open-source projects at  https://github.com/octoberzzzzz/ML-based-TMS-CAV-Survey.
\end{abstract}

%\begin{IEEEkeywords}
%{Trust management system, connected autonomous vehicles, supervised machine learning, semi-supervised machine learning, reinforcement learning}
%\end{IEEEkeywords}


\IEEEpeerreviewmaketitle

% === I. INTRODUCTION =============================================================
% =================================================================================
\section{Introduction}

\IEEEPARstart {C}{onnectivity}, intelligence, electrification and sharing are key trends reshaping the automotive industry. Connected Autonomous Vehicles (CAVs) are products of highly connected and autonomous levels, transforming traditional transportation means into mobile agents equipped with powerful sensory, computational and data storage capabilities. CAVs promise to revolutionize transportation by improving efficiency, enhancing safety, and enabling sustainable operations. However, cyber threats have grown significantly over the past decade, characterized by large-scale, high-frequency attacks with a high degree of invisibility~\textcolor{blue}{\cite{upstream2023global}}. In response, 
regulatory frameworks such as UN R155~\textcolor{blue}{\cite{UNR155}}(the first mandatory vehicle cybersecurity regulation) and ISO/SAE 21434~\textcolor{blue}{\cite{ISO21434}} (the first international vehicle cybersecurity standard) have been established since 2021. 

These regulations have accelerated CAVs and Intelligent Transportation Systems (ITS) into a new era, shifting focus from basic safety to integrated safety-security-privacy frameworks. The defense-in-depth principle has been implemented in vehicular security through a two-tier framework, where proactive defenses against certificate-based internal threats (Line-1) and responsive countermeasures against external threats (Line-2) are employed. While a rule-based Intrusion Detection System (IDS) detects attacks through predefined signatures or anomaly thresholds, it fails to recognize malicious behaviors by entities with legitimate identities. In contrast, the trust mechanism serves as a cross-layer framework, facilitating prevention and real-time protection. The trust mechanism is expected to ensure CAVs evolve from basic intelligence to verifiable trust.

Trust is defined as a set of activities and a security policy, where element x trusts element y if and only if x has confidence that y will behave in a well-defined way that does not violate the given policy~\textcolor{blue}{\cite{isoiec27036-1}}. The Trust Management System (TMS) is a way of organizing the trust mechanism. TMS can be classified into trust value-based TMS, trust value-free TMS (i.e.,cryptography-based solutions~\textcolor{blue}{\cite{Asaar2018}},~\textcolor{blue}{\cite{Adil2022}}), and hybrid TMS. 
This survey focuses on trust value-based TMS, which quantifies trust through a series of steps.

\subsection{Motivations of ML-based TMS for CAV}
Traditional TMS methods rely on statistical weighting, inference modeling and graph theory. In contrast, machine learning (ML)-as a branch of artificial intelligence(AI)-offers a data-driven approach to uncovering patterns, making predictions and automating decisions. The growing adoption of ML-based TMS for CAV (TMS-CAV) has been driven by four key factors.

\begin{itemize}
    \item \textbf{CAV intelligence and massive multi-source heterogeneous data}: The generation of multi-source data enables ML applications across CAV perception, decision-making, and control. Notable advances include: ML-based collaborative perception, autonomous driving decision with deep reinforcement learning (DRL), and traffic prediction with Graph Neural Networks (GNN).

    \item \textbf{Cross-domain intelligence for TMS}: ML-based TMS solutions have succeeded in Internet of Things (IoT), social networks and other domains. These methodologies provide valuable insights for adapting TMS to CAV-specific challenges.

    \item \textbf{Theoretical innovations driven by CAV characteristics}: The high dynamism, real-time demands, and collaborative nature of CAVs require novel TMS frameworks. Traditional methods (e.g., Bayesian Interference and Fuzzy Logic) struggle with inaccuracy and inefficiency. Instead, modern ML methods like GNNs can better capture dynamic CAV network topologies.

    \item \textbf{Advances in ML methods}: Breakthroughs in ML, such as Transformer, DRL, and Federated Learning, demonstrate strong potential to address TMS-CAV challenges, including privacy preservation and real-time processing.
\end{itemize}

\subsection{Scope and a Brief Comparison with Current Surveys}

Despite the increasing number of studies on ML-based TMS-CAVs, few studies have systematically reviewed this critical intersection.
To bridge this gap, this survey synthesizes existing work on ML-based TMS-CAVs and provides forward-looking insights through innovative analytical perspectives. To the best of our knowledge, this survey represents the first in-depth survey of ML-based TMS-CAVs, with a particular focus on the latest advancements in both ML and TMS research.

Given the rapid evolution of TMS, CAVs, and ML, comprehensive coverage is impractical. Consequently, this survey adopts an exploratory approach, including emphasizing core literature, critically analyzing current studies and emerging challenges, and pointing out future research directions. Methodologically, while traditional Systematic Literature Reviews (SLRs) typically follow the PRISMA (Preferred Reporting Items for Systematic Reviews and Meta-Analyses) framework, this survey implements a simplified PRISMA process to enhance transparency and standardization.

\begin{figure}
  \begin{center}
  \includegraphics[width=3in]{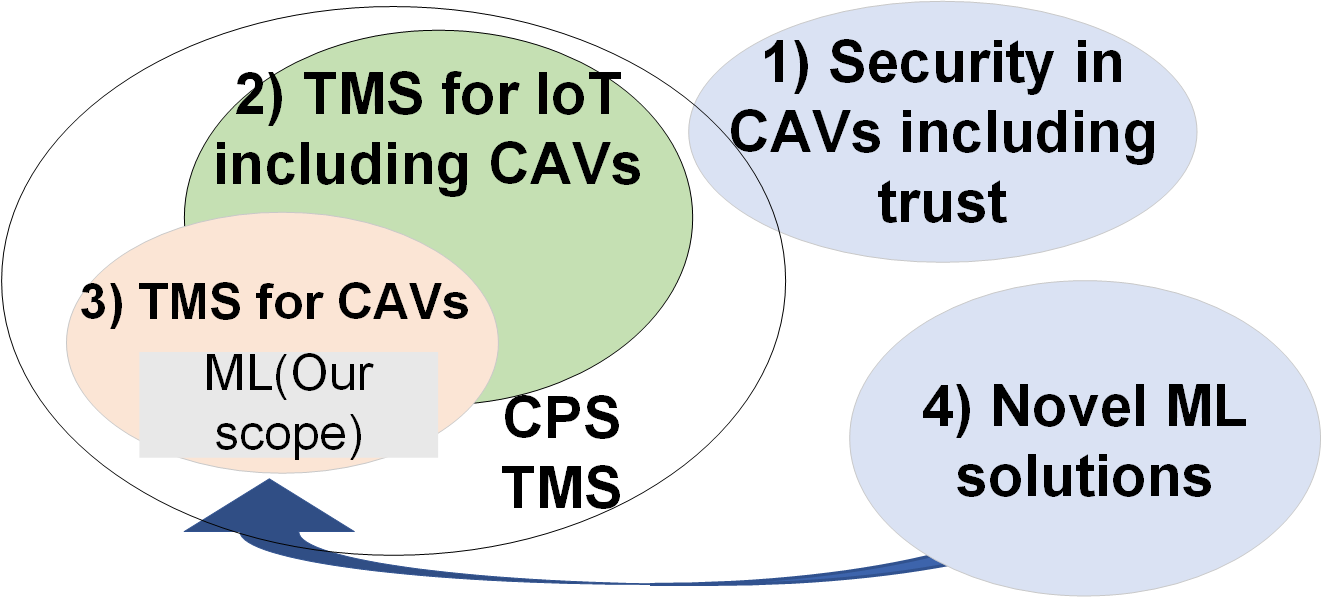}\\
  \caption{Scope of the survey}
  \label{fig:scope}
  \end{center}
\end{figure}

As shown in~\textcolor{blue}{Fig.\ref{fig:scope}}, prior surveys are categorized into five overlapping yet distinct categories. A detailed comparison of representative surveys is provided in Section II; here, we highlight key distinctions. The scope of this survey is more narrowly focused but offers a more in-depth exploration.

\begin{itemize}
    \item \textbf{Different themes:} Existing surveys on security in CAVs (Type 1) treat trust as a subset of broader security frameworks. Meanwhile, surveys on TMS for IoTs (Type 2) focus on generic trust solutions with CAVs as one of many applications. Surveys on TMS-CAVs (Type 3) examine trust methods for CAVs, but often relegate ML methods to subsidiary sections. Surveys on novel ML solutions (Type 4) summarize the ML advances with less focus on TMS or CAV. Therefore, surveys on ML-based TMS for CAVs (Type 5) uniquely bridge ML-based TMS and CAV-specific challenges.
    
    \item \textbf{Innovative perspectives:}  Unlike prior TMS-CAV surveys (Type-3), this survey of Type-5 distinguishes itself through several innovative perspectives. Rather than simply cataloging methods, a novel taxonomy is proposed based on the hierarchical architecture of TMS. Additionally, we investigate how ML enhances each functional module.
    
    \item \textbf{Diverse CAV application scenarios:} Beyond cataloging methods, this survey emphasizes practical applications of TMS in CAV and ITS contexts, demonstrating their real-world efficacy.
    
\end{itemize}

% =======
% FIG. 01
% =======

\subsection{Challenges of Surveying on ML-based TMS for CAVs}
This survey identifies four key challenges in current research, spanning classification methods, theoretical foundations, CAV applications, and ML techniques.

Challenges-1 {\textbf{Lack of a novel taxonomy for ML-based TMS for CAVs:} Existing surveys (Type 1-4) are tailored to their specific topics and research scopes. Traditional TMS methods and ML-based TMS methods differ fundamentally in their theoretical foundations and classification criteria. While ML employs taxonomies like supervised or unsupervised learning, TMS involves several processes from trust-related data collection, trust computation to trust-related attack defense. Thus, it poses a challenge for this survey, where existing taxonomies are insufficient to cover emerging ML-based TMS and CAV needs.

Challenges-2 \textbf{Lack of systematic theoretical foundation and evaluation criteria:}
While existing TMS surveys (Type 2) have established theoretical frameworks and evaluation criteria for IoT and social networks, these prove inadequate for ML-based TMS in CAVs. Traditional TMS methods differ fundamentally from ML-based TMS, particularly in CAV contexts. Moreover, the conventional criteria focusing solely on effectiveness prove inadequate for evaluating ML-based TMS for CAVs. New criteria incorporating ML-specific metrics (precision, recall, F1-score) are needed to properly assess performance in CAV contexts.

Challenges-3 \textbf{Insufficient integration of TMS with CAV applications:} 
TMS methods have been employed in advanced techniques, such as edge computing and crowdsensing. Meanwhile, these techniques are also important in ITS and CAVs. However, existing surveys on TMS for IoTs (Type 2) and TMS for CAVs (Type 3) 
mainly adopt methodological approaches as taxonomy criteria, rather than CAV applications. This gap highlights the urgent need to establish a novel taxonomy of CAV application-driven TMS.}

Challenges-4 \textbf{Limited coverage of advanced ML methods:} ML methods continue to advance in key areas, including privacy-preserving techniques, model interpretability, and defense against adversarial attacks. These solutions are also essential for ML-based TMS for CAVs. Recent surveys on novel ML solutions (Type-4) have documented these advancements on ML, and some studies have begun to utilize them. However, existing surveys of TMS for IoTs (Type-2) and TM for CAVs (Type-3) have failed to sufficiently cover these advanced ML methods.

\subsection{Contributions}
To address the aforementioned challenges and meet the requirements of an exploratory survey, this survey has four key contributions:
% =======
% FIG. 01
% =======

\begin{figure}
  \begin{center}
  \includegraphics[width=3.6in]{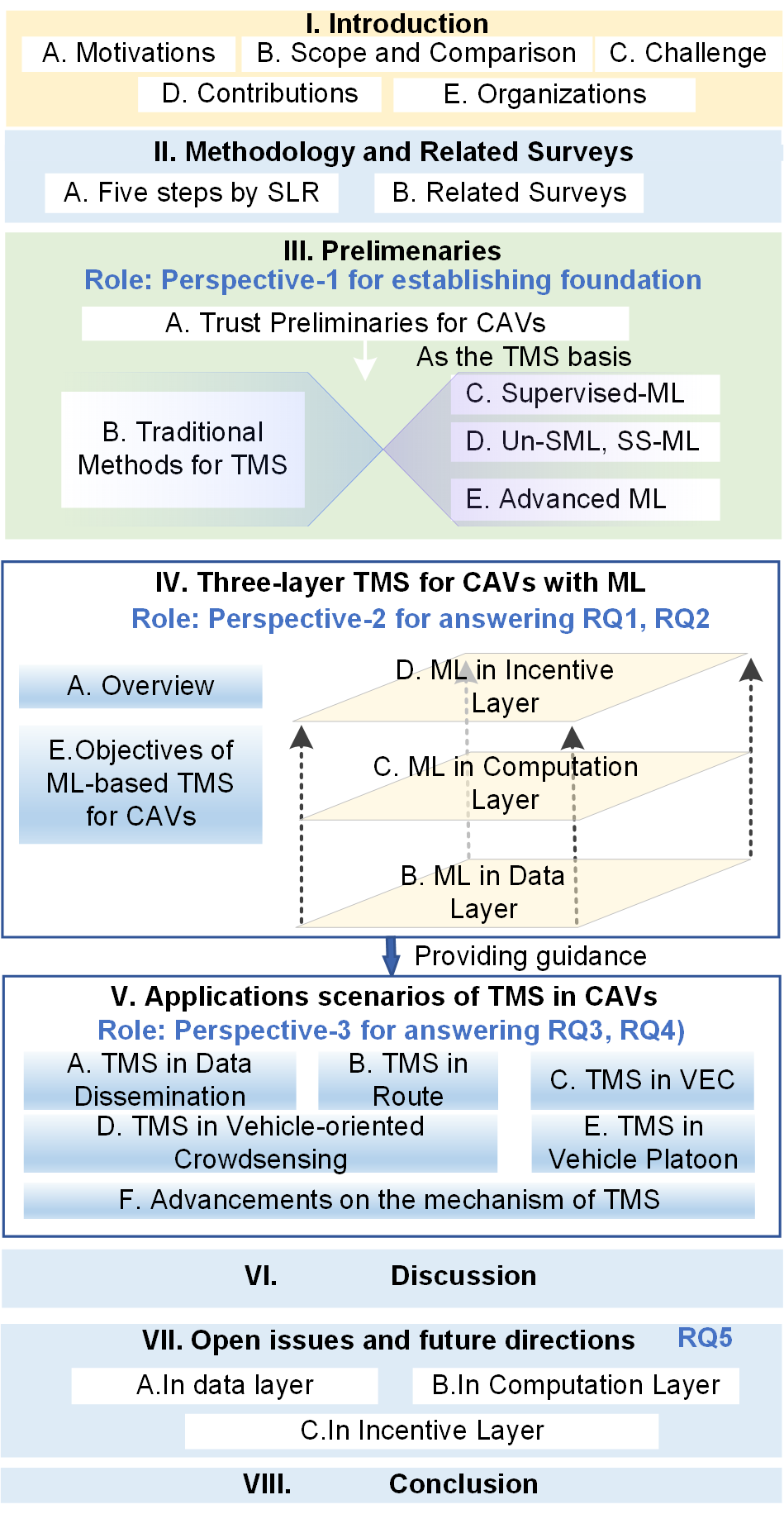}\\
  \caption{Organization framework of this survey}
  \label{fig:org}
  \end{center}
\end{figure}

\subsubsection{\textbf{Establishing a three-layer ML-based TMS framework for CAVs assisting the description of the survey (Addressing Challenge-1)}} 
A novel three-layer framework for ML-based TMS in CAVs is established, comprising
the trust data layer, the trust computation layer and the trust incentive layer. Unlike existing TMS surveys for IoTs (Type-2) and CAVs (Type-3) that focus on methods and individual steps (mainly ML-based trust assessment), a more systematic perspective is provided where ML applications are considered across all TMS components.

\subsubsection{\textbf{Summarizing the principle and evaluation criteria of ML-based TMS for CAVs (Addressing Challenge-2)}} The requirements of ML-based TMS for CAVs are clarified. Suitable ML methods for each level are analyzed and compared with traditional TMS methods. The advantages and challenges of ML-based TMS for CAVs are analyzed. Comprehensive evaluation metrics are developed by combining standard ML performance metrics (e.g., F1-score) with CAV-specific requirements like real-time processing capabilities and system scalability.

\subsubsection{\textbf{Evaluating the latest ML-based TMS methods according to common CAV scenarios (Addressing Challenge-3)}} It is more practical to discuss under CAV scenarios. Given that it is still in its early stage, traditional methods are also referred in the same scenarios, as well as relevant advances in other areas. We summarize the commonalities and differences of TMS for CAVs in different scenarios.

\subsubsection{\textbf{Summarizing work on advanced ML (Addressing Challenge-4)}}
Beyond basic ML methods, the advanced ML methods are also considered, such as DRL, GNN and federated learning. Their potential applications on TMS are also analyzed. Open issues and future directions are subsequently analyzed.

\subsection{Organizations}
As shown in~\textcolor{blue}{Fig.\ref{fig:org}}, the survey is organized as follows:
Section II explains the methodology of this survey and compares it with related surveys.
Section III explains the concept of trust and examines both traditional methods and ML algorithms for TMS, highlighting their applications and limitations.
Section IV proposes and analyzes the three levels of TMS for CAVs, discussing the opportunities and challenges of ML and proposing evaluation metrics.
Section V reviews the latest works on TMS in various CAV scenarios, with a focus on ML-based solutions.
Section VI provides a discussion of the current state of research and identifies gaps.
Section VII outlines open issues and future directions.
Section VIII makes a conclusion.

\section{Methodology and Related Surveys}
\subsection{Five Steps by the PRISMA Method}
The following five key steps are used to ensure transparency and reproducibility. Core literature and critical analysis are emphasized.

\subsubsection{Formulation of the research question(RQ)}
Based on the analysis of the challenges and opportunities, this survey is organized around the following five RQs.

RQ1: What are the benefits and mechanics of applying ML methods to every step of TMSs for CAV? It aligns with Contribution 1) and Contribution 2).

RQ2: What are the criteria for ML-based TMS for CAVs? It aligns with Contribution 2).

RQ3: Do the current works fulfill these criteria? It aligns with Contribution 2).

RQ4: How can TMS be combined with emerging ITS applications to guarantee better services? It aligns with Contribution 3).

RQ5: What are the future directions for improving existing ML-based TMSs in CAVs? It aligns with Contribution 4).

\subsubsection{Search strategy}

Based on these RQs, many combinations of keywords are proposed for the search engine. Specifically, four groups of keywords are listed:
Group I includes keywords about trust, trust management and TMS.
Group II includes keywords related to ML.
Group III includes keywords related to CAVs.
Group IV includes keywords related to the purpose of TMS, especially security.

The following combinations are used to identify specific types of studies.
i) Group III + Group V = Studies on Security in CAVs. ii) Group I + Group IV = Studies on TMS for IoTs.
iii) Group I + Group III = Studies on TMS for CAVs.
iv) Group II = Studies on security in novel ML solutions.
v) Group I + Group IV + Group III + Group V= Studies on ML-based TMS for IoT.
vi) Group I + Group IV + Group III + Group V= Studies on ML-based TMS for CAVs.

After searching and distinguishing, specific publications have been placed in the appendix for reproduction. As shown in {Fig.\ref{fig:journaltheme}}, the growth trend of publication volume over time is given. Publications of TMS for CAVs are rapidly increasing, inspired by TMS for IoT. Thus, surveying the ML-based TMS for CAVs is significant.

\begin{table}[h!]
\centering
\caption{Categories and keywords}
\label{tab:keywords}
\begin{tabular}{cl}
\toprule
\textbf{Categories} & \textbf{Keywords} \\
\midrule
Group I & trust management, trust model, trust aggregation, \\
        & trust computation, trust assessment, reputation \\
\addlinespace
Group II & machine learning, deep learning, artificial intelligence, \\
         & Reinforcement Learning, federated learning, \\
         & Graph Neural Network \\
\addlinespace
Group III & Connected and Automated Vehicles, Internet of Vehicles, \\
          & V2V, V2X \\
\addlinespace
Group IV & IoT \\
\addlinespace
Group V & security, trust, privacy \\
\bottomrule
\end{tabular}
\end{table}

\begin{figure}
  \begin{center}
  \includegraphics[width=3.6in]{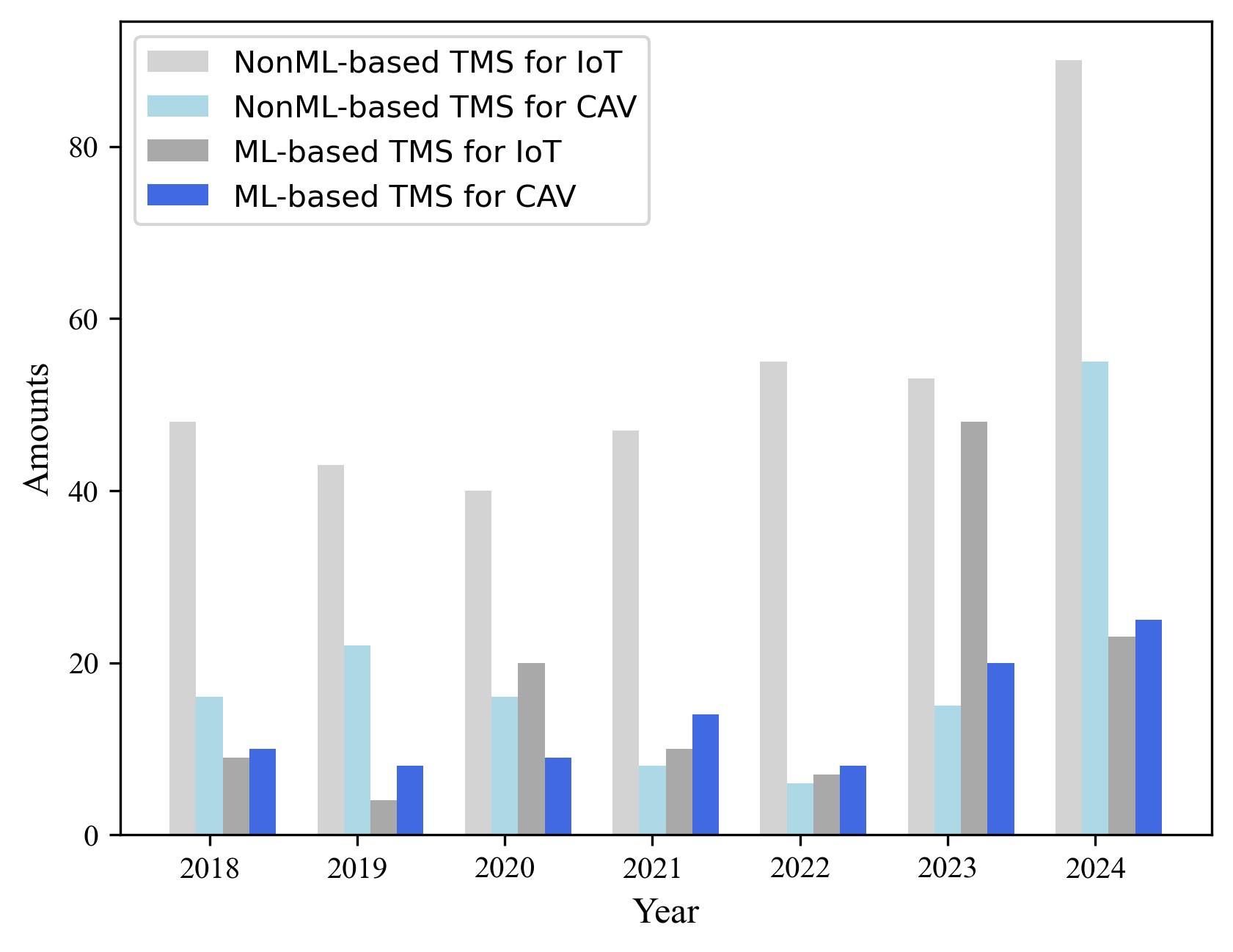}\\
  \caption{Growth trend of publication volume over time}
  \label{fig:TMSyear}
  \end{center}
\end{figure}

\subsubsection{Rigorous screening process}
References were selected based on quality, methodological novelty, trust indicators, and proposed techniques. Specifically, the inclusion criteria are as follows.

\begin{itemize}
    \item Language and type: Papers written in English and in the Web of Science core index.
    \item Contents: Keywords in Table I should appear in the title, abstract, or keywords. 
    \item Range: From 2018 to the present.
    \item Topic Priorities: i) ML-based TMS for CAVs. ii) ML-based TMS methods for IoT (due to the limited number of suitable studies currently available, though this is increasing). iii) ML methods for ITS. iv) Traditional TMS methods for CAVs, which serve as a comparative baseline.
    \item Manuscript Priorities: i) Highly cited regular papers (20+ citations) from top-tier journals (JCR Q1/Q2) from 2018 to present. ii) Recent studies in journals from 2022 to present. iii) High-quality surveys and studies on the conference from 2018 to the present.
\end{itemize}   

   The exclusion criteria are as follows:
\begin{itemize}
       \item ML methods not related to CAV or TMS.
       \item References to ML-based TMS lacking sufficient detail.
       \item Studies not peer-reviewed.
       \item Studies with methodological flaws, unreliable data, or lower citation before 2020. 
       \item Hybrid encryption-based trust and trust value-based studies are considered, but studies only covering encryption-based trust are excluded.
       \item Some ML-based TMS methods for OSN are considered, but studies covering only social trust with traditional methods are excluded. 
\end{itemize}

\begin{figure}
  \begin{center}
  \includegraphics[width=3in]{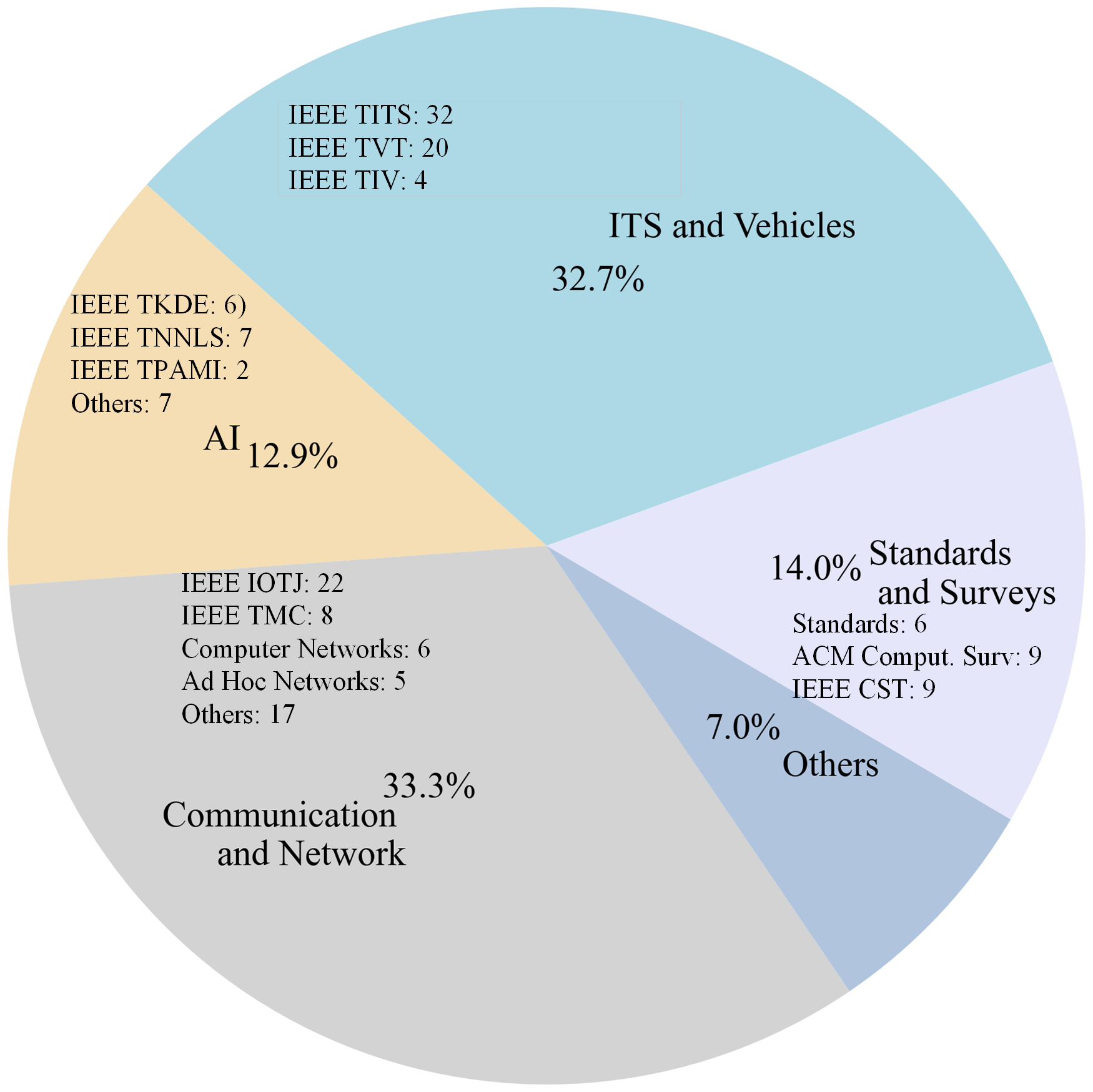}\\
  \caption{Distribution of included publications by journal source after screening}
  \label{fig:journaltheme}
  \end{center}
\end{figure}

Due to space limitations that prevent the inclusion of all surveyed publications in the reference, the number of publications is 172 that meet rigorous selection criteria. We investigated their journal source. As shown in {Fig.\ref{fig:journaltheme}}, this distribution analysis demonstrates the breadth of authoritative sources underpinning our survey and identifies dominant publication venues in this research domain. The statistical process will be shown in our repository\footnotemark{}.
\footnotetext{https://github.com/octoberzzzzz/ML-TMS-CAV-Survey}.

\subsubsection{Information extraction and analysis}
The following information on existing research is identified and summarized, i.e., date, type, strengths, weaknesses, research gaps, and potential research directions.

\subsubsection{Quality assessment and critical analysis}
Qualitative and quantitative methods are used to answer the above five RQs.

\subsection{Related Surveys}
Aligned with the five-category classification of existing surveys outlined in Section I, we systematically examine the first four types. This survey on ML-based TMS for CAVs fills a critical gap as the fifth category, for which no prior survey exists.

\subsubsection{Surveys on security in CAVs including trust}
Tracing the trajectory of CAV security research, the survey by ~\textcolor{blue}{\cite{van2019}} in 2019 pioneered systematic investigations of misbehavior detection in C-ITS and laid conceptual groundwork for TMSs through its comparative analysis of node-centric versus collaborative architectures. 
In 2019, security, privacy and trust of Vehicular Ad-Hoc Networks (VANET) were discussed together in~\textcolor{blue}{\cite{Lu2019}}. Recent surveys also highlight these issues, such as security for autonomous vehicles using ML and blockchain\textcolor{blue}{\cite{Bendiab2023}} and V2X cybersecurity \textcolor{blue}{\cite{Sedar2023}}, both published in 2023.

These surveys systematically summarized CAV security methods and introduced TMS concepts for CAV, but they did not provide a detailed summary of TMS methods for CAV. 

\subsubsection{Surveys on TMS for IoT}
A large number of TMS on the different types of IoT are proposed, such as edge-based IoT~\textcolor{blue}{\cite{Liu2023BlockTMS}}, blockchain edge-based IoT~\textcolor{blue}{\cite{Fotia2023edge}}, 
social IoT~\textcolor{blue}{\cite{sagar2024social}}, the general IoT~\textcolor{blue}{\cite{Wei2022}}, heterogeneous network~\textcolor{blue}{\cite{Wang2022HeTrust}}. In 2022, a taxonomy of trust models~\textcolor{blue}{\cite{Wang2022HeTrust}} is proposed, consisting of decision models, evaluation models and management models, where ML methods are briefly discussed in part of the evaluation models. 

These surveys provide a solid understanding of the basics of TMS, including trust attributes, components of trust management, systems organizations beyond simple trust assessments, and classifications of TMS. However, they lack a detailed discussion on CAV and more advanced ML methods. 

\subsubsection{Surveys on TMS for CAVs}
The subsequent research has extended and started to specialize in TMS for CAVs. In 2022, Hussain et al.,~\textcolor{blue}{\cite{Hussain2022}} conducted a comprehensive survey on TMS for VANET, detailing the features of VANETs impacting TMS in VANET, methods for TMS in VANET, and TMS in enabling techniques. In 2022, Hbaieb et al.,~\textcolor{blue}{\cite{hbaieb2022IoV}} analyzed 
V2X challenges, including security issues posed by V2X characteristics, trust-related attacks, and security attacks for IoV. In 2022, Zavvos et al.,~\textcolor{blue}{\cite{Zavvos2022}} proposed a taxonomy of privacy and trust for IoV at the service level. 

These surveys examined TMS for CAVs from a broad perspective, and their coverage was limited by focusing primarily on studies from 2014 to 2022, when ML-based TMS approaches were relatively scarce. They have identified future research directions in ML-based TMS and trust-based services. Instead of ML-based TMS as a branch, we conduct a specialized, in-depth examination of ML-based TMS for CAV.

\subsubsection{Surveys on Novel ML solutions in security solutions and ITS}
Macas et al.,~\textcolor{blue}{\cite{macas2022survey}} surveyed the deep learning applied in cybersecurity, but TMS methods were not covered. Luo et al.,~\textcolor{blue}{\cite{luo2021deep}} surveyed deep learning methods in anomaly detection for Cyber-physical Systems (CPS) but did not mention TMS. Boualouache et al.,~\textcolor{blue}{\cite{Boualouache2023}} proposed a taxonomy of ML-based misbehavior detection systems for 5G and beyond vehicular networks. More ML methods are applied to ITS, such as federated learning~\textcolor{blue}{\cite{Zhang2024Fed}},~\textcolor{blue}{\cite{Chellapandi2024Fed}}, graph-neural networks~\textcolor{blue}{\cite{li2024graph}}, and transfer learning, etc. 

These surveys analyzed the application of novel ML in ITS and recognized the importance of trust in open challenges. However, TMS methods were not combined with the transportation scenarios.

\subsubsection{Surveys on ML-based TMS}
In 2020, Wang et al., surveyed the ML-based trust evaluation~\textcolor{blue}{\cite{wang2020survey}}. It only discussed traditional ML methods and did not consider the application of GNN, RL, and other advanced ML methods. Subsequently, with the rapid development of these methods, Luo et al.,~\textcolor{blue}{\cite{LuoGNNTrust2025}} summarized the GNN-based trust evaluation in 2025.

Our survey on ML-based TMS for CAV is of this type. It extends prior general ML-based trust assessment by focusing on CAV applications and advancing from trust evaluation to comprehensive steps of TMS. Besides, it incorporates recent progress across TMS, ML and CAV.

\section{Preliminaries of Trust and Machine Learning}
% =======
% FIG. 02
% =======
\begin{figure*}
  \begin{center}
  \includegraphics[width=6in]{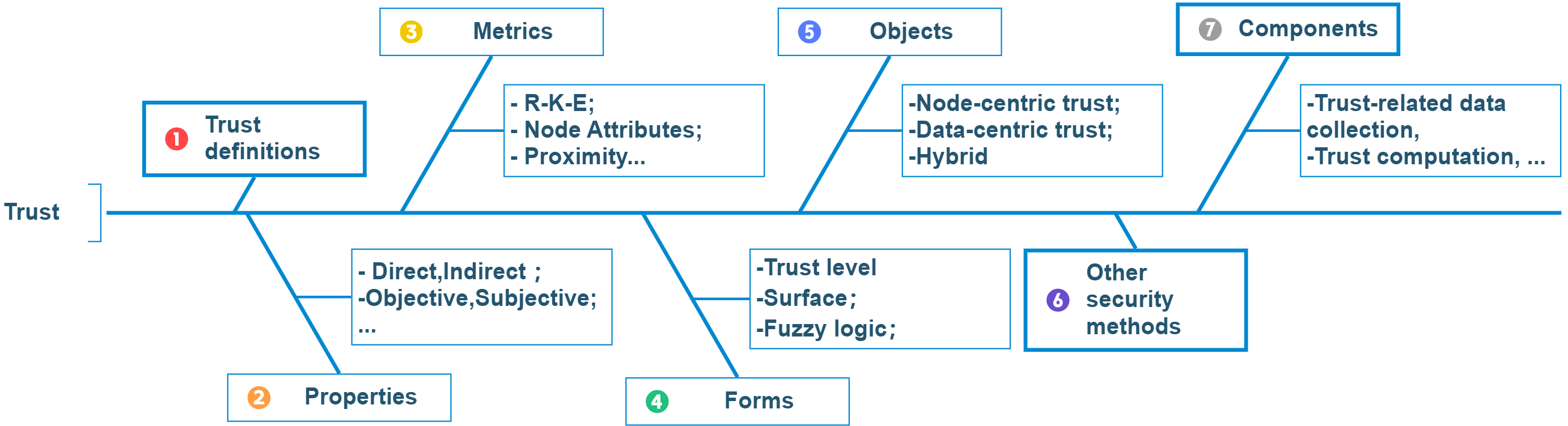}\\
  \caption{Framework of trust preliminaries}
  \label{fig:pre}
  \end{center}
\end{figure*}

While previous surveys have systematically discussed the preliminaries of trust and ML, our work emphasizes those concepts that are particularly relevant to CAVs and their integration with ML methods.

\subsection{Trust Preliminaries for CAVs}
In~\textcolor{blue}{Fig.\ref{fig:pre}}, seven parts are to build trust preliminaries.

\subsubsection{Current various trust definitions}

(1) Trust in IEC 27036-1~\textcolor{blue}{\cite{BSISOIEC27036-1}} and GBT/25069~\textcolor{blue}{\cite{GBT25069}}: the relationship between two entities and/or elements, consisting of a set of activities and a security policy in which element x trusts element y if and only if x has confidence that y will behave in a well-defined way (concerning the activities) that does not violate the given security policy. 

(2) Trust in various network types: 
CPS can be seen as an advanced form of IoT. IoT focuses on the connection between “things”. Internet of Everything (IoE) is an extension of IoT that also encompasses the full connectivity of people, processes, data and things. Automation of Everything (AoE) focuses on automating various processes and tasks.

In physical and cyber domains of CPS, trust is the expectation that a physical or cyber entity will complete a task as intended, as shown in~\textcolor{blue}{\cite{ITUY3052}}. Trust influences an entity's preference for services and its decision to engage in transactions with others. Trust studies are important in IoT, IoE, AoE, and CPS, where the scope and complexity are increasing.

(3) Trust in various stakeholders of autonomous vehicles (AV):
While AV focuses on the autonomous driving capabilities of the vehicle itself, CAV emphasizes the connectivity and collaboration of vehicles.
As shown in~\textcolor{blue}{Table \ref{tab:stakeholder_trust}},
different stakeholders of AV have different trust focuses. This delineates the specific beneficiaries of trust mechanisms.

\begin{table}[h]
\centering
\caption{Trust in various stakeholders in AVs}
\label{tab:stakeholder_trust}
\begin{tabular}{ll}
\toprule
\textbf{Stakeholder} & \textbf{Trust focus} \\ 
\midrule
City authorities & public safety, traffic efficiency \\ 
Policymakers & liability definition, data privacy \\ 
Police & law enforcement compatibility, data forensics \\ 
Manufacturers & technical redundancy, AI transparency \\ 
Users & safety, sense of control \\ 
Pedestrians & behavior predictability, interaction clarity \\ 
\bottomrule
\end{tabular}
\end{table}

(4) Trust in various AV types by application purpose: The features of six AV types are compared in {Table \ref{tab:av_classification}. 
Thus, as demonstrated in Table \ref{tab:AVtype_trust}},
trust focuses vary significantly in various AV types. These AVs are not necessarily CAVs. It is essential to recognize that distinct service scenarios and user demographics necessitate tailored optimization strategies.

\begin{table*}[h!]
\centering
\caption{Classification of AV by application purpose}
\label{tab:av_classification}
\renewcommand{\arraystretch}{1.2}
\begin{tabularx}{\textwidth}{l*{7}{>{\centering\arraybackslash}X}}
\toprule
\multirow{2}{*}{\textbf{Classification Dimension}} & \multicolumn{7}{c}{\textbf{Autonomous Vehicle Types}} \\
\cmidrule(lr){2-8}
 & \textbf{Logistics AV} & \textbf{Public Transit AV} & \textbf{Robotaxi} & \textbf{FMLM AV} & \textbf{Shared AV} & \textbf{Fleet AV} & \textbf{Delivery AV} \\
\midrule

\textbf{Core Function} 
& Goods transport & Mass transit & On-demand mobility & Short-haul transfer & Time-sharing lease & Urban capacity & Contactless delivery \\

\textbf{Operating Speed (km/h)} 
& 80--100 & 30--50 & 40--60 & <20 & 40--70 & 25--45 & 15--30 \\

\textbf{Road Type} 
& Highway/Freight & Dedicated BRT & Urban roads & Community paths & Mixed roads & Fixed routes & Sidewalk-compatible \\

\textbf{ODD Complexity} 
& Medium & Low & \cellcolor{yellow!25}Very High & Medium & High & Low & Medium \\

\textbf{V2X Requirement} 
& $\bigstar$$\bigstar$$\bigstar$$\frac{1}{2}$ 
& $\bigstar$$\bigstar$$\bigstar$$\bigstar$ 
& $\bigstar$$\bigstar$$\bigstar$$\bigstar$$\frac{1}{2}$ 
& $\bigstar$$\bigstar$$\bigstar$$\frac{1}{2}$ 
& $\bigstar$$\bigstar$$\bigstar$$\bigstar$$\frac{1}{2}$ 
& $\bigstar$$\bigstar$$\bigstar$$\bigstar$$\bigstar$$\frac{1}{2}$ 
& $\bigstar$$\frac{1}{2}$$\frac{1}{2}$ \\

\textbf{Policy Barriers} 
& Low & \cellcolor{red!15}Very High & High & Low & Medium & High & Medium \\

\bottomrule
\end{tabularx}

\vspace{2mm}
\footnotesize \textit{Note:} ODD = Operational Design Domain; V2X ratings based on SAE J3161 standards. Star symbols represent relative requirement levels.
\end{table*}

\begin{table}[h]
\centering
\caption{Trust in various AV types}
\label{tab:AVtype_trust}
\begin{tabularx}{\linewidth}{>{\raggedright\arraybackslash}p{3cm}X}
\toprule
\textbf{AV Types} & \textbf{Trust focus} \\
\midrule
Logistics AVs & System robustness, reliability, and cybersecurity \\
Public transit AVs & System robustness, safety, and passenger comfort \\
Robotaxi & Multi-user privacy protection, safety, and reliability \\
First mile/last mile (FMLM) AVs & Accessibility, operational flexibility, and pedestrian interaction safety\\
Shared self-driving vehicles & Data sharing transparency, multi-user coordination, and hygiene management\\
Public autonomous transit fleet & System scalability, emergency handling, and schedule reliability\\
Fully autonomous delivery & Package security, contactless operation, and urban adaptability\\
\bottomrule
\end{tabularx}
\end{table}

(5) Trust from three layers: i) At the technical layer, efforts should be made to enhance the reliability and transparency of CAVs, ensuring their safety and stability in complex environments. ii) At the policy and regulatory layer, strict testing standards and safety regulations need to be established to provide a solid legal foundation for the commercial application of CAVs. iii) At the user layer, extensive public awareness campaigns, test drive experiences, and other initiatives should be conducted to improve public understanding and acceptance of CAV technology. Currently, stakeholders are actively taking action in these three layers.

\subsubsection{Trust definitions in CAV in this survey} 
It is defined by referring to the current various trust definitions.

(1) Standards and layers: The concepts are consistent with the standards.This survey focuses on the progress and challenges at the technical level.

(2) Network types: Current studies on trust in IoV are regarded as network-oriented trust, which concerns the IoV environment or the trust relationships between IoV participants, i.e., how nodes trust each other and securely exchange information. As IoV evolves towards more integrated CPS, IoE, and AoE frameworks, the scope of trust will likely expand to include the trust of the entire connected ecosystem, from the physical environment to the digital infrastructure and the automated processes that support them.

Thus, trust in CAV in this survey focuses on agent-oriented trust and service-oriented trust based on network-oriented trust, i.e., trust should serve for specific CAV scenarios, such as platoon leader vehicle selection, trustworthy edge devices selection to offload computations, etc. It is from AoE and beyond the scope of IoT and IoV. Part of the research on trust in IoV covers how edge nodes mutually trust, but the object of trust in CAV is centralized on vehicles.
    
(3) Stakeholders: This survey mainly analyzes trust in CAVs from a technical perspective. Here, city authorities, policymakers, and law enforcement are the regulators of trust; users and pedestrians are the beneficiaries of trust; and manufacturers are the builders and trustees of trust.

(4) Distinguishing trust in TMS from credibility: Credibility in positioning~\textcolor{blue}{\cite{Xu2024CrePos}} focuses on the accuracy and consistency of results and system performance in complex environments. Instead, trust in TMS focuses on the trust of interacting entities and its impact on decisions and behavior.

(5) Smart city technologies enhance ML-based TMS for CAVs: Advanced techniques in smart cities enable more efficient ITS. Digital twin technology, combined with AI, improves CAV trust through precise decision-making. Blockchain also supports ML-based TMS for CAVs.
    
(6) Vehicle types in terms of application purpose: A generalized set of ML-based TMS for CAVs with extension modules for different vehicle types in {Table \ref{tab:av_classification} should be proposed. Compared with the trust of single-vehicle intelligence, the TMS of CAV is more complex because it not only relies on the performance of the vehicle itself, but also relies on various factors such as IoV and multi-subject coordination.}

(7) Vehicle types in terms of capabilities: CAVs represent a higher level of Grade of Automation (GoA) and the cooperative perception capabilities. Administrators and leading companies are making significant progress toward achieving GoA-5 (full automation) and Level-3 (highest cooperative perception). As a result, the development and integration of advanced TMS are essential as a technical preparation to support the efficient and reliable operation of CAVs. 

(8) Vehicle types in terms of penetration rate: Current research on TMS for CAVs 
predominantly examines idealized scenarios with 100\% CAV penetration rates, neglecting the more realistic mixed traffic environment. Since current autonomous technologies cannot handle 100\% scenarios independently, they must achieve an 'acceptable safety threshold' through human-machine collaboration, rather than pursuing the impractical goal of 'absolute safety'. Since current autonomous technologies cannot handle all operational scenarios independently, they must achieve an acceptable safety threshold through human-machine collaboration, rather than pursuing the impractical goal of absolute safety. Human-in-the-Loop (HITL) and Human-on-the-Loop (HOTL) architectures have emerged as critical components for establishing trust in CAV systems. Therefore, developing robust TMS-CAV frameworks is essential for ensuring reliable and secure operations in mixed traffic environments.
    
\subsubsection{Properties of trust in CAV}
Several properties are analyzed in pairs, such as direct and indirect trust. Additionally, trust is pseudo-asymmetry, historical relevance, context dependence, and dynamic, etc.

(1) Direct, Indirect and Hybrid: Most TMS for CAV literature consider these properties. Direct trust comes from traffic conditions observed by on-board sensors or IoV, while indirect trust relies on recommendations from neighboring vehicles or RSUs when direct data is insufficient.

(2) Objective and Subjective: 
Objective trust is derived from measurable and quantifiable data, while subjective trust involves personal or collective perceptions, experiences, or emotions.
Objective trust is the primary focus in CAV application.

(3) Local and Global: Local trust is evaluated over a limited geographical range and short periods, making it suitable for real-time decision-making. Global trust spans broader regions and longer durations, providing a more comprehensive but less immediate assessment.

(4) Pseudo-asymmetry: It means one entity may trust another without reciprocity. 
The pseudonyms mechanism of CAV complicates trust evaluation by requiring repeated trust computations for the same entity under different identities.

(5) Historically relevant and context-dependent: Trust is influenced by both past behavior and contextual conditions, such as traffic conditions changes.

(6) Social: In social IoV, trust may incorporate social attributes, such as community-based reputation.

(7) Time-evolving: Trust is dynamic and continuously updated in response to new data, environmental shifts, or behavioral changes.

\subsubsection{Metrics of Trust}
Properties of trust describe how trust is established, while metrics of trust quantify and analyze trust through specific criteria. They are divided into reputation-based, knowledge-based, expectation-based, node properties-based,proximity-based and environment-based~\textcolor{blue}{\cite{hbaieb2022IoV}}. Reputation, knowledge and proximity-based factors are the most widely used metrics~\textcolor{blue}{\cite{hbaieb2022IoV}}. ML-based TMS methods automatically extract features from data, prioritizing sufficiently high-quality inputs. Unlike traditional methods requiring expert-designed metrics, ML models need not predefine fixed trust criteria. However, metrics remain valuable for enhancing the explainability of ML.

\subsubsection{Forms of trust}
\begin{figure}[h]
    \centering
    \begin{subfigure}[b]{0.23\textwidth}
        \includegraphics[width=\textwidth]{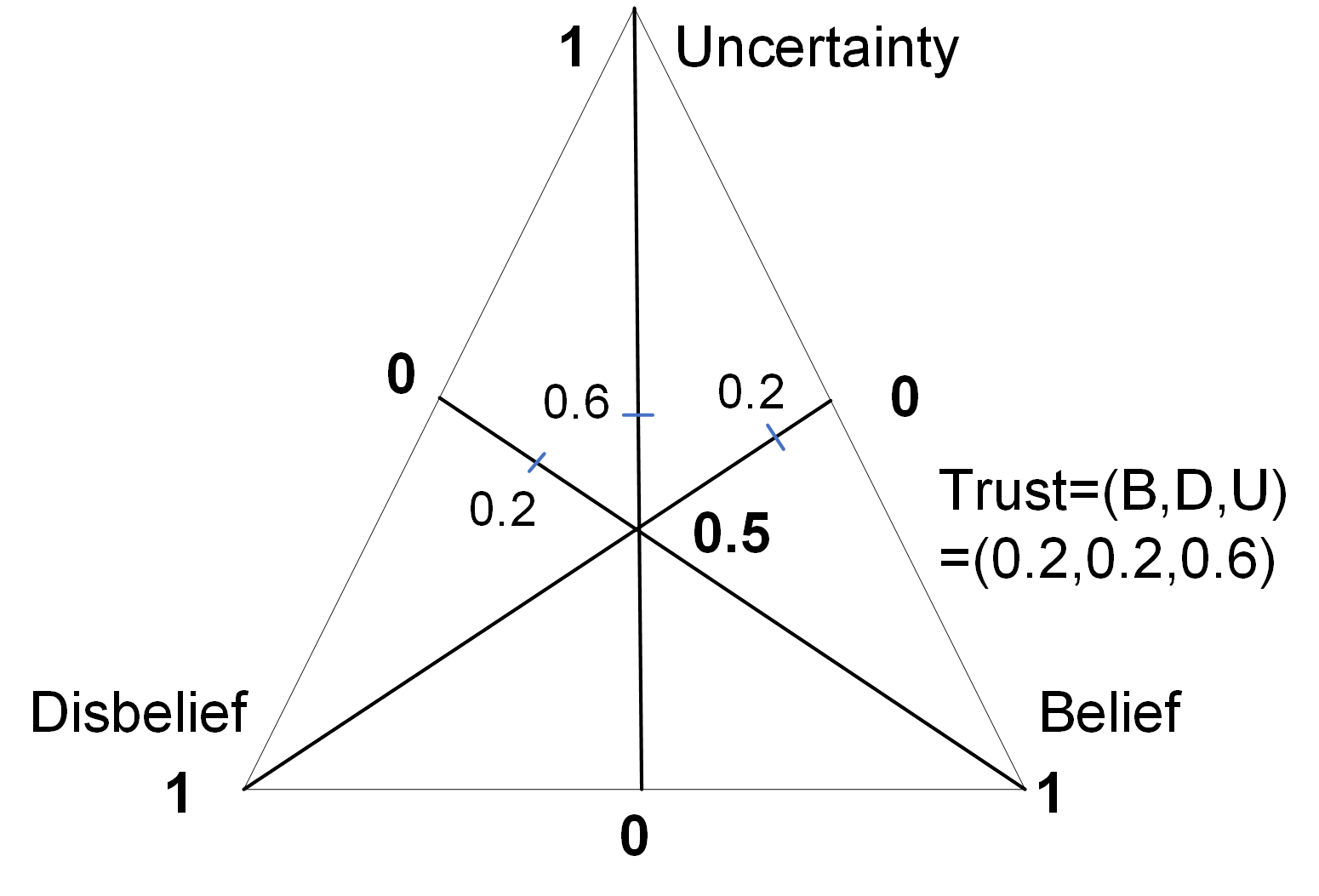}
        \caption{trust facets class}
        \label{fig:con_pos}
    \end{subfigure}
    \hspace{-0.1cm} % 调整水平空隙
    \begin{subfigure}[b]{0.23\textwidth}
        \includegraphics[width=\textwidth]{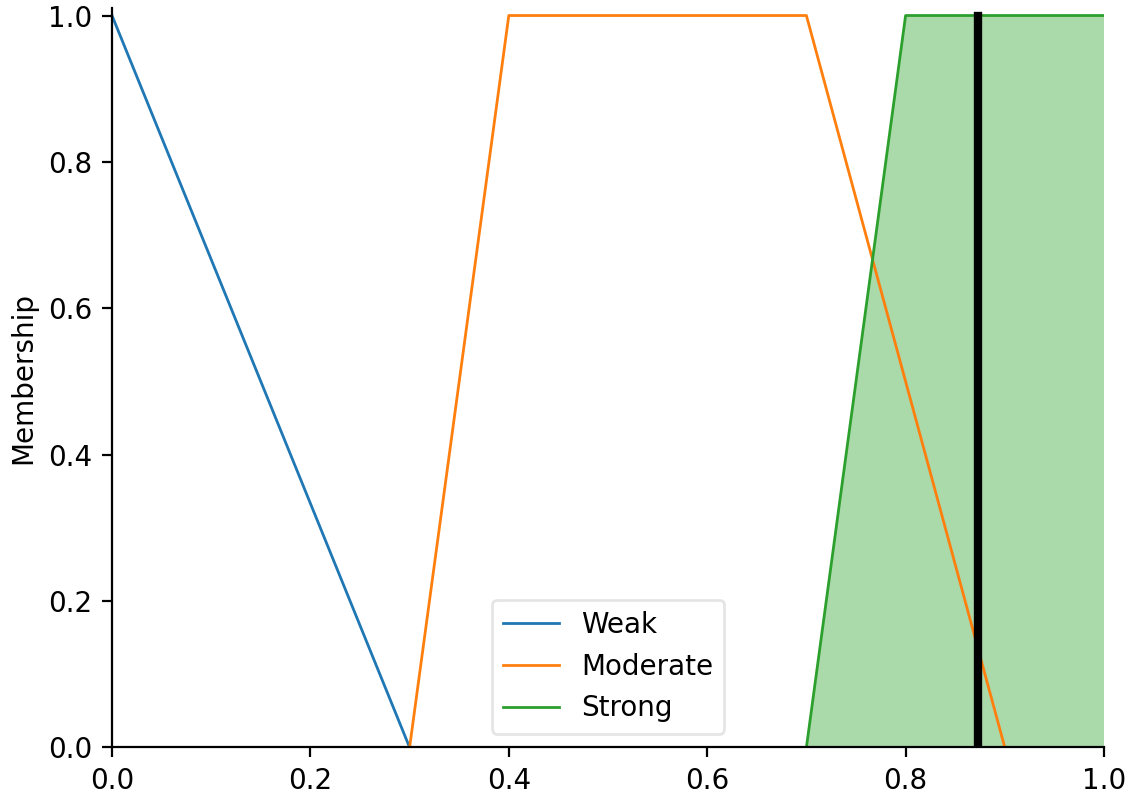}
        \caption{trust logic class(Partial process from~\textcolor{blue}{\cite{Hasan2023Fuzzy}})}
        \label{fig:dos_pos}
    \end{subfigure}
    \hspace{-0.1cm} % 调整水平空隙
    \begin{subfigure}[b]{0.25\textwidth}
        \includegraphics[width=\textwidth]{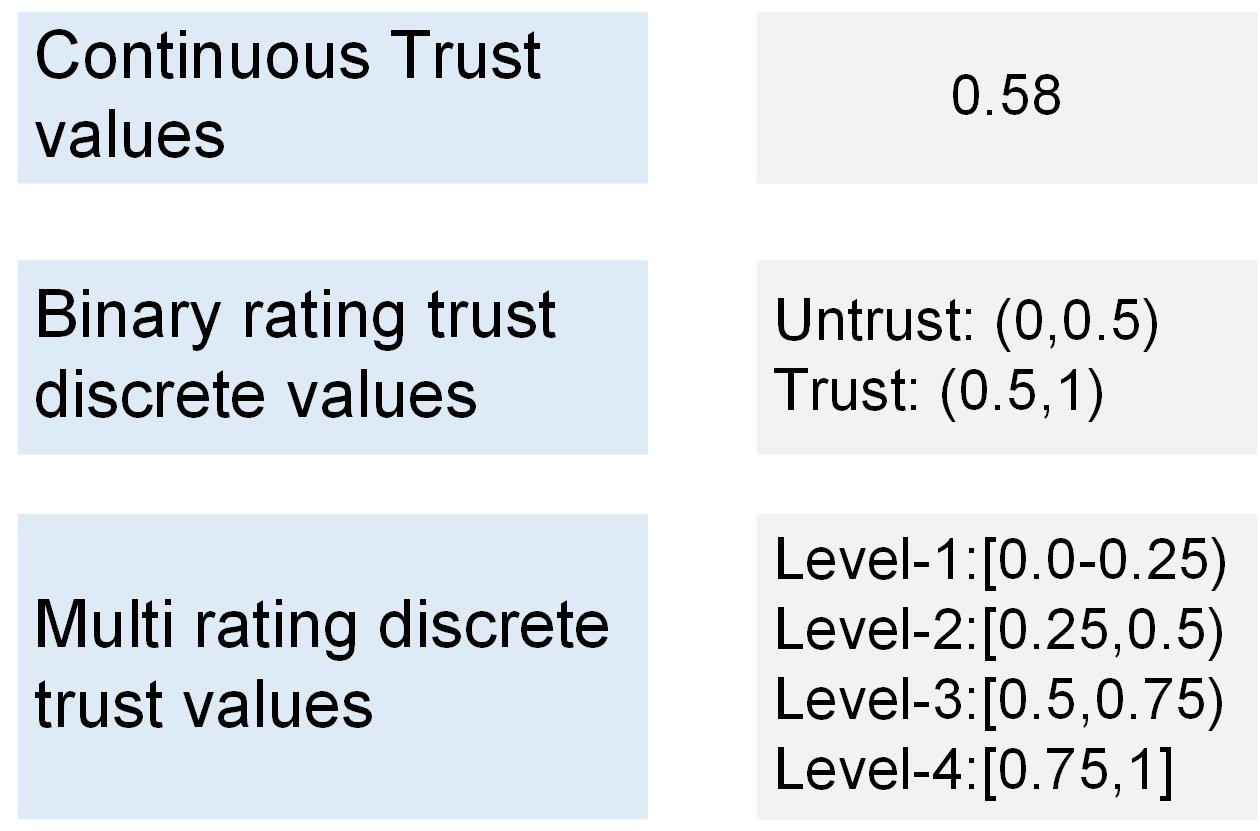}
        \caption{trust level class}
        \label{fig:gridsybil_pos}
    \end{subfigure}
    \caption{Three forms of trust}
    \label{fig:form}
\end{figure}

They vary across methods as shown in~\textcolor{blue}{Fig.\ref{fig:form}}. We categorize forms as follows referred by~\textcolor{blue}{\cite{hbaieb2022IoV}}:
\begin{itemize}
    \item Non-ML-based Trust: i) Trust facets: Represented by tuples (e.g., $(B,D,U)$ for Belief, Disbelief, Uncertainty in D-S evidence theory) ii) Trust logic: Expressed through probabilities or fuzzy values. 
    \item ML-based Trust: Trust levels are described by continuous or discrete values. i) Continuous values (Regression): [0,1] range where 1 means highest trust. ii) Binary rating (ML binary classification): trust-1, distrust-0. iii) Multi-rating (ML multi-class classification): The values in [0,1] are divided into several levels. 
    
\end{itemize}

\subsubsection{Objects of trust}
i) Data-centric: Evaluates data trust (e.g., emergency message dissemination). ii) Node-centric: Assesses node or entity (e.g., trusted edge node selection). iii) Hybrid: Combines both data and node trust evaluation.

\subsubsection{TMS differences from other security methods}
Unlike IDS, TMS requires more data and multi-step processing. While TMS can enhance IDS by detecting misbehaviors~\textcolor{blue}{\cite{Li2022TrustIDS}}, its trust value also supports broader decision-making applications, such as edge computing offloading and platoon formation.

\subsubsection{Components of TMS}
These terms fall under trust value-based TMS, including trust-related data collection, trust computation/assessment/evaluation, trust propagation, trust aggregation, trust prediction and trust data storage~\textcolor{blue}{\cite{Hussain2022}}. 

\subsection{Traditional Methods for TMS}
\begin{figure*}
  \begin{center}
  \includegraphics[width=6in]{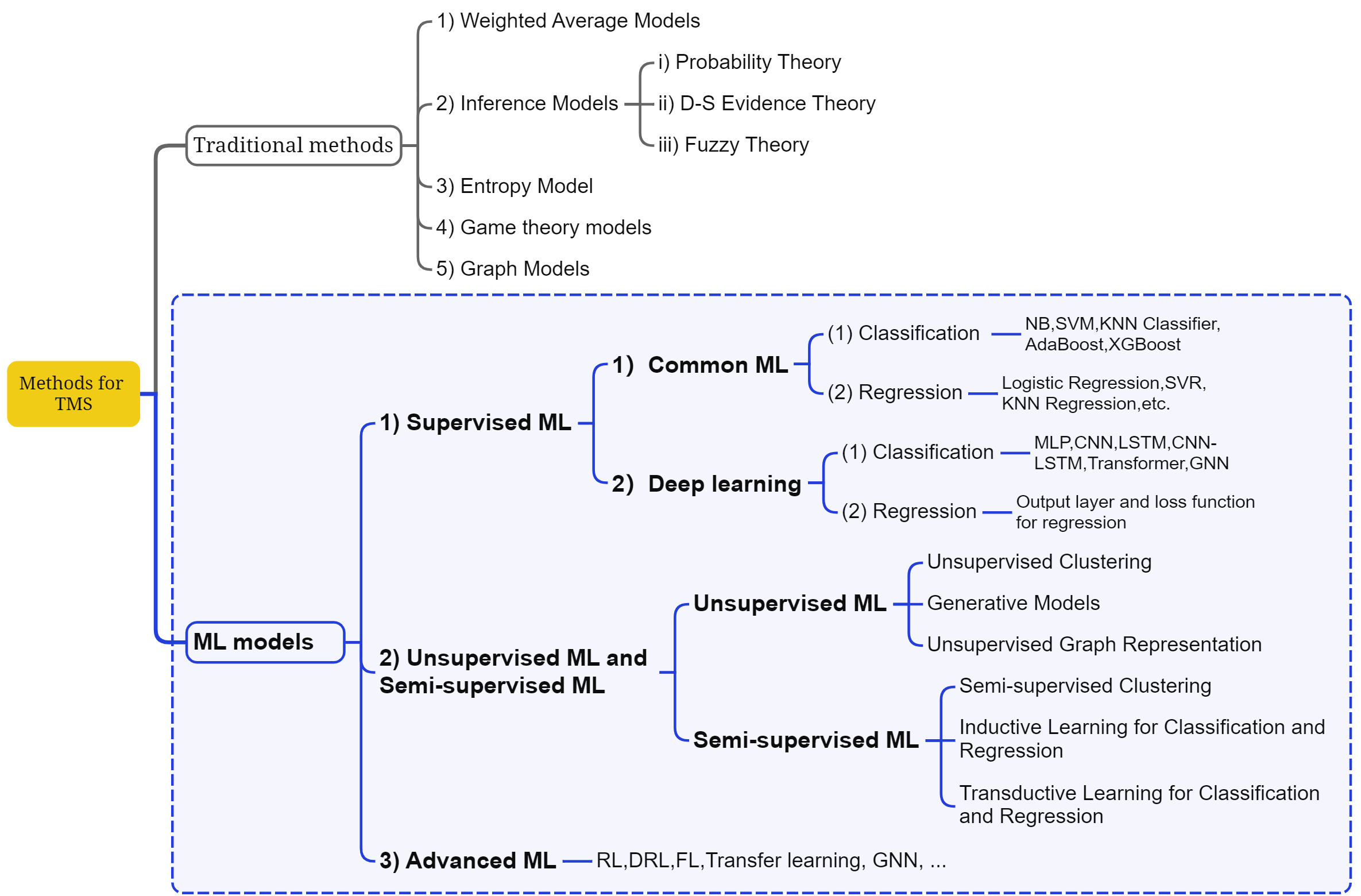}\\
  \caption{Methods for TMS}
  \label{fig:method}
  \end{center}
\end{figure*}

Methods for TMS are shown in~\textcolor{blue}{Fig.\ref{fig:method}}, traditional TMS methods are briefly introduced as the basis.

% =======
% FIG. 04
% =======

\subsubsection{Statistically weighted TMS}
This most widely used approach weights multi-source information for comprehensive trust evaluation. 
Applications include secure content delivery~\textcolor{blue}{\cite{Li2020Incentive}}, routing~\textcolor{blue}{\cite{sho2023tgrv}}, cluster head selection~\textcolor{blue}{\cite{kadam2023CH}}.

\subsubsection{Inference-based TMS} 
i) \textit{Probability-based TMS}:  Trust is quantified as a probability value. Applications include the beta-Bayesian model for malicious CAV node detection~\textcolor{blue}{\cite{Gao2022}}. 
 ii) \textit{D-S evidence-based TMS:} Trust is defined by combining and processing the conflicting evidence to quantify uncertainty as confidence degrees~\textcolor{blue}{\cite{Zhang2023DSTSurvey}}. It is suitable for uncertainty or prior-knowledge-lacking scenarios. Applications include fusing multi-dimensional trust metrics for CAV~\textcolor{blue}{\cite{Cheong2024MalNode}}.
iii) \textit{Fuzzy logic-based TMS:} Rather than providing a definite value, it estimates the trust degree using rules and an approximate set. Applications include fuzzy evaluation~\textcolor{blue}{\cite{Hasan2023Fuzzy}}, ~\textcolor{blue}{\cite{Zhao2022EntitySocial}} and ~\textcolor{blue}{\cite{Shao2023SDN}}; the interval type-2 fuzzy logic~\textcolor{blue}{\cite{Yang2023GAN}}. 

iii) \textit{Subjective logic-based TMS:} 
Subjective logic handles uncertainty and subjectivity in TMS. Applications include Three-valued Subjective Logic (3VSL) for CAV~\textcolor{blue}{~\cite{Cheng2019SubLog}}, subjective logic~\textcolor{blue}{~\cite{Mirzadeh2023MalRoute}}.

\subsubsection{Entropy-based TMS}
It evaluates the randomness of information, signals or events~\textcolor{blue}{~\cite{Guo2020EntropyV_TMS}}.

\subsubsection{Graph-based TMS}
When graphs represent entities and their relationships, graph algorithms can be used for TMS, like OpinionWalk~\textcolor{blue}{~\cite{Cheng2019SubLog}} and RandomWalk for TMS~\textcolor{blue}{~\cite{Dhelim2023Trust2vec}}.

\subsubsection{Game-based TMS}
It seeks to identify and eliminate selfish nodes through a game, such as evolutionary game theory for TMS~\textcolor{blue}{~\cite{Tian2019EvoGame}}.

\subsection{ML Methods: Supervised vs. Unsupervised vs. Semi-Supervised Learning}

Based on different data labeling situations and learning objectives, ML methods are categorized by supervised learning (as Eq.1), unsupervised learning, and semi-supervised learning. First of all, supervised learning aims to train a model \( f \) that maps inputs \( x \) to labels \( y \) by minimizing a loss function on labeled data.

\begin{equation}
\min_{f \in \mathcal{F}} \sum_{i=1}^{N} L(y_i, f(x_i))
\end{equation}

\begin{itemize}
    \item $f \in \mathcal{F}$: Model from hypothesis space $\mathcal{F}$.
    \item $\mathbf{x}_i$: Input feature vector (training example $i$).
    \item $y_i$: True label for $\mathbf{x}_i$.
    \item $L$: Loss function comparing $f(\mathbf{x}_i)$ and $y_i$.
    \item $N$: Number of training examples.
\end{itemize}

\subsubsection{Supervised Traditional ML Methods for Classification}

These ML methods deal with linear or non-linear relationships in data but do not involve complex hierarchical structures.

(1) \textit{Originally designed for binary-class classification but can be extended to multiple-class classifications.} This category includes: Naive Bayes (NB) Classifier, Support Vector Machines (SVM), and Adaptive Boosting (AdaBoost).
Notably, AdaBoost-SAMME supports multi-class classification.

(2) \textit{Multiple-class classifications.} This category includes: K Nearest Neighbours (KNN) classifier, Random Forests classifier and eXtreme Gradient Boosting (XGBoost).

\subsubsection{Supervised Deep Neural Networks for Classification}
(1) Classic deep learning methods.
Deep learning excels at processing large-scale datasets with high-dimensional features. For binary classification tasks, binary cross-entropy loss functions are typically employed, whereas standard cross-entropy loss functions are used for multi-class classification problems. Classical models are Multilayer Perceptron (MLP), Convolutional Neural Network (CNN),
Recurrent Neural Network (RNN) like Long Short-Term Memory (LSTM) and Gated Recurrent Unit (GRU).

(2) Emerging models.
i) Hybrid Architectures: The CNN-LSTM model integrates CNNs for spatial feature extraction with LSTM for sequential modeling, making it particularly suitable for spatio-temporal data analysis.

ii) Transformer: Characterized by self-attention mechanisms, multi-head attention, and residual connections, Transformer models efficiently capture long-range dependencies. Their parallel processing capability enables faster training than traditional RNN architectures.

iii) GNN: GNNs specialize in processing graph-structured data through node and edge relationships, supporting four primary classification tasks: node classification,
edge classification, graph classification
link prediction.
Key supervised GNN variants include Graph Convolutional Networks (GCNs), Graph Attention Networks (GAT), and GraphSAGE. Spatiotemporal GNNs (STGNNs) further enhance modeling capabilities by combining graph networks with temporal learning methods, demonstrating strong performance in traffic flow prediction and trajectory forecasting applications~\textcolor{blue}{\cite{Jin2023STGNN}}.

\subsubsection{Supervised ML Methods for Regression}
These methods can predict continuous trust values and can be categorized as:

(1) Traditional ML regression: It includes Logistic Regression (LR), Random Forest Regression, Support Vector Regression (SVR), and K-Nearest Neighbors Regression (KNN Regression).

(2) Deep learning regression: It differs from classification in three key aspects:
i) Output layer uses linear or constant activation (vs. softmax or sigmoid)
ii) Loss functions use Mean Squared Error (MSE) or Root Mean Squared Error (RMSE), instead of cross-entropy.
iii) Evaluation metrics include continuous error measures instead of classification metrics.

\subsubsection{Unsupervised ML Methods}
Unsupervised learning aims to train a model \( f \) that discovers patterns or structure in unlabeled data \( x \) by optimizing an objective function (e.g., clustering, density estimation)

\begin{equation}
\min_{f \in \mathcal{F}} \sum_{i=1}^{N} \mathcal{D}(x_i, f(x_i))
\end{equation}

\begin{itemize}
    \item $\mathcal{D}(\cdot)$ denotes a dissimilarity measure (e.g., reconstruction error, KL divergence).
    
    \item $f(x_i)$ may correspond to cluster assignments, latent embeddings, or synthetic samples, depending on the task.
\end{itemize}
 
The following categorization proofs are also referred from recent surveys, including self-supervised learning~\textcolor{blue}{\cite{Li2023Selfsupervised}},~\textcolor{blue}{\cite{Gui2024SelfSuper}}; graph-based semi-supervised learning~\textcolor{blue}{\cite{Song2023GraphSemi}}; semi-supervised learning to cyber-security~\textcolor{blue}{\cite{mvula2024Semi}}; semi-supervised graph clustering~\textcolor{blue}{\cite{dan2024semisurvey}}.
It is divided as follows:

(1) Unsupervised clustering: 
It groups similar data points into distinct clusters while maximizing dissimilarity between clusters. Common methods include K-means, hierarchical clustering, Density-Based Spatial Clustering of Applications with Noise (DBSCAN), spectral clustering, and Gaussian mixture models.

(2) Generative models: They learn data distributions to synthesize new instances, including Generative Adversarial Network (GAN), Variational Autoencoder (VAE), diffusion models, autoregressive networks, and normalizing flows.

(3) Unsupervised graph representation: a) Graph Contrastive Learning: These methods learn a node or graph representation by maximizing the similarity of positive sample pairs in the graph and minimizing the similarity of negative sample pairs.  Common methods are Deep Graph Infomax (DGI), Graph Contrastive Learning with Augmentations (GCA), InfoGraph, etc. 
b) Graph Generative Learning: This method applies a generative model to graph data to learn the structure and properties of the graph and generate new graph instances or graph elements (e.g., nodes or edges). Common methods are GAE, VGAE, Graph GANs, etc. 

\subsubsection{Semi-supervised ML Methods} 
It trains a model $ f $ to map inputs $ x $ to labels $ y $ by jointly leveraging labeled data (via supervised loss) and unlabeled data (via regularization or pseudo-labeling).

\begin{equation}
\min_{f \in \mathcal{F}} \underbrace{\sum_{i=1}^{L} L(y_i, f(x_i))}_{\text{Supervised loss}} + \lambda \cdot \underbrace{\sum_{j=L+1}^{N} \mathcal{R}(x_j, f)}_{\text{Unlabeled data term}}
\end{equation}

where:
\begin{itemize}
    \item $ L $ is the number of labeled samples, $ N $ is the total number of samples.
    \item $ \mathcal{R}(\cdot) $ regularizes using unlabeled data (e.g., consistency loss, entropy minimization).
    \item $ \lambda $ balances the two terms.
\end{itemize}

(1) Semi-supervised Clustering: Clustering does not involve the prediction of labels but explores the intrinsic structure of the data. Semi-supervised clustering algorithms use limited labeled data to improve clustering results. It is divided as follows: a) Metric-based. b) Partial-labels.c) Density-based. d) Constraint-based. e) Hierarchical clustering. f) Outcome variable association. g) Model-based clustering. h) Graph-based clustering.

(2) Inductive Learning for Classification and Regression: These methods aim to learn models that can be generalized to new data, which improves the model's ability to predict new data by learning embedded representations of nodes or using neural networks to capture the global data structure.
\begin{itemize}
    \item Wrappers: It includes self-training methods, co-training methods and boosting methods.
    
    \item Unsupervised Preprocessing: It includes a) Cluster-then-label. b) Pre-training. c) Feature extraction.
    \item Intrinsic SSL: It includes a) Max-margin. b) Perturbation-based. c) Generative model: d) Manifolds.
    \item Graph-based: It includes a) Graph-based label propagation methods. b) Sparse model. c) Low-rank model. d) GNN. 
    
\end{itemize}

(3) Transductive Learning for Classification and Regression: These methods focus on predicting labels for specific unlabeled nodes in the current dataset and do not emphasize the generalization capabilities of the model. These methods typically use local information in the graph structure to propagate labels and are suitable for situations where the dataset is relatively fixed. It includes embedding-based label propagation and label propagation based on GNN.

\subsection{Advanced ML Methods}

\subsubsection{RL, DRL and MADRL}
RL enables agents to learn optimal behaviors through environmental interactions. DRL enhances RL with deep learning for complex tasks~\textcolor{blue}{\cite{Wang2024DRL}}. MADRL extends this to multi-agent systems. 

There are many classification criteria for RL, e.g., value-based RL and policy-based RL. Since trust calculation methods based on 
RL methods are categorized by:
\begin{itemize}
    \item Discrete action space: Q-Learning, SARSA, DQN, Policy Gradient, Actor-Critic, Proximal Policy Optimization (PPO), MADQN, etc. 
    \item Continuous action space: Deep Deterministic Policy Gradient (DDPG), Asynchronous Advantage Actor-Critic (A3C), Trust Region Policy Optimization (TPRO), MADDPG, MA-Actor-Critic. Besides, the variance methods of DQN are DDQN, D3QN, etc. 
    \item Research directions: the combination of GNN and RL~\textcolor{blue}{\cite{Munikoti2023DRLGNN}}, offline RL or online RL~\textcolor{blue}{\cite{Prud2023OffRL}}, safe RL for safety-critical scenarios~\textcolor{blue}{\cite{mo2024secuRL}}, transfer learning with RL~\textcolor{blue}{\cite{Zhu2023TransferDRL}}.
\end{itemize}

\subsubsection{Federated Learning}
In FL, edge devices or clients only send the gradients or the learnable parameters to cloud servers rather than sending massive local datasets in a centralized learning framework. FL is one of the edge intelligence algorithms. FL has many classification criteria, such as by the topology of the participating devices in the FL system, including centralized FL, decentralized FL, hierarchical FL and clustered FL~\textcolor{blue}{\cite{chen2023FL}}.
FL provides new training and testing for ITS tasks, such as trajectory prediction, traffic prediction, V2X, etc~\textcolor{blue}{\cite{Zhang2024Fed}}. FL is closely combined with blockchain~\textcolor{blue}{\cite{zhu2023blockchainFL}} and edge computing~\textcolor{blue}{\cite{Duan2023FLEC}}.

\subsubsection{Other Novel ML Methods}
\begin{itemize}
    \item Transfer learning: 
It leverages knowledge from solved problems to address related new tasks. It can be integrated with RL and FL~\textcolor{blue}{\cite{Zhu2023TransferDRL}}.
   \item Incremental learning (or continual learning): It enables models to continuously learn from new data streams. Yang et al.~\textcolor{blue}{\cite{Yang2024FLCon}} developed federated continual learning through knowledge fusion.
   
   \item Meta learning: It enables rapid task adaptation by transferring prior knowledge (meta-data), eliminating full retraining~\textcolor{blue}{\cite{gharoun2024meta}}.
\end{itemize}

\subsection{Summary on Potential ML Methods for ML-based TMS-CAV}
After surveying the mechanisms, categories, and recent advances of ML, the following insights are obtained. Additionally, more insights should be further integrated with TMS functional modules and CAV-specific operational scenarios.
\subsubsection{Advantages of Semi-Supervised Learning}
Supervised ML-based TMS requires extensive labeled data, which is challenging to obtain for dynamic CAV environments. Semi-supervised methods address this by effectively leveraging limited labeled data alongside abundant unlabeled data, outperforming purely unsupervised approaches.
\subsubsection{Advantages of Advanced ML}
While traditional ML initially dominated CAV-TMS, modern techniques now offer superior solutions, such as Transformers, GNNs, RL, FL, transfer learning, etc.. For example, Transfer learning shows particular promise in mitigating cold-start issues, as demonstrated in EV charging forecasting~\textcolor{blue}{\cite{Forootani2024TransferCold}}, though its application to TMS-CAV remains rare.

\subsubsection{Advantages of Appendix Methods}
In order to avoid performance degradation or failure of ML-based TMS methods, auxiliary measures for TMS should be considered, such as integration with IDS and cryptographic methods.

\section{Three-layer Trust Management System for CAVs Enhancing by Machine Learning}
\subsection{Three-layer TMS for CAV}
\begin{figure*}
  \begin{center}
  \includegraphics[width=5in]{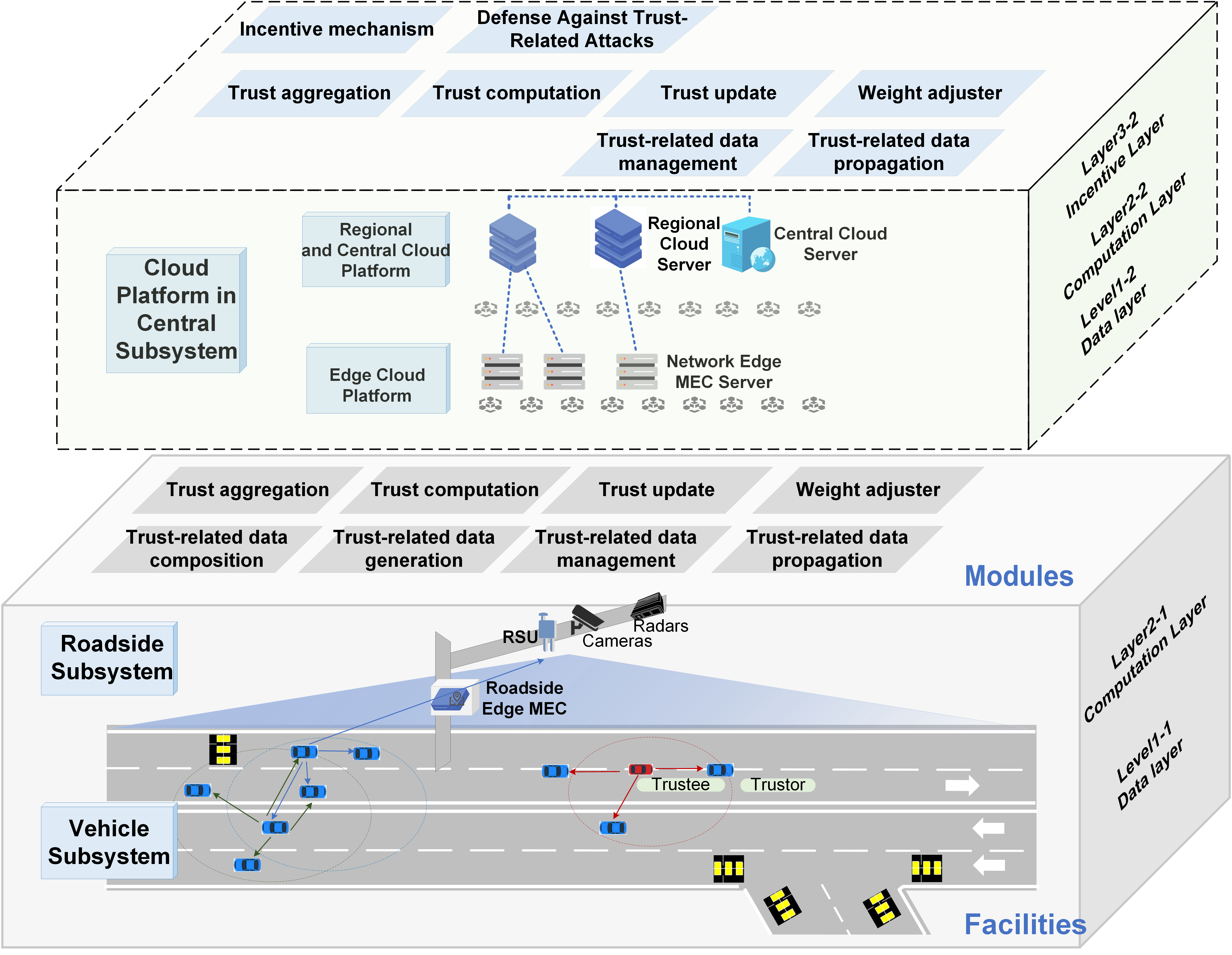}\\
  \caption{CAV in Vehicle Road-Cloud Integration System}
  \label{fig:FrameCAV}
  \end{center}
\end{figure*}

To facilitate our survey in describing the functional modules of the TMS and how ML can enhance the role of each module, a three-layer ML-based TMS for CAV is proposed. As shown in~\textcolor{blue}{Fig.\ref{fig:FrameCAV}}, it adapts IoT-based TMS principles to vehicle-road-cloud systems. The modules in each layer maintain consistency with TMS in IoT from~\textcolor{blue}{\cite{Wei2022}} while integrating ML methods for CAV environments. It offers a holistic perspective that surpasses prior surveys focused on isolated modules.

\begin{enumerate}
    \item Trust Data Layer: It defines CAV what data can be regarded as trust-related data and the data analysis process. It consists of four modules: i) Trust-related Data Composition, ii) Trust-related Data Generation, iii) Trust-related Data Management and iv) Trust-related Data Dissemination. 
    \item Trust Computation Layer: It obtains trust values from trust-related data. It consists of four modules. i) Trust Aggregation; ii) Trust Calculation; iii) Trust Update; and iv) Weight Adjuster.
    \item Trust Incentive Layer: It solves
    passive and aggressive behaviors to guarantee reliability and robustness. 
    It consists of two modules. i) Incentives; ii) Trust-related Attacks Defense. 
\end{enumerate}

% =======
% FIG. 
% =======

As illustrated in ~\textcolor{blue}{Fig.\ref{fig:FrameCAV}}, the initial computation of both the trust data layer and the trust computation is deployed at the vehicle subsystem and the roadside subsystem. 
Deeper processing across all three layers is allocated to central subsystems due to their substantial resource requirements. 
Additionally, the ML requirements for TMS must account for CAV features that differentiate them from social IoT, WSN, and UASNs implementations, as also highlighted in the survey on TMS for VANET~\textcolor{blue}{\cite{Hussain2022}}. Specifically, 

\begin{itemize}
\item \textbf{Speed:} CAV nodes move at varying speeds, causing the network topology to change constantly, with no permanent neighbors. Therefore, ML-based TMS methods designed for CAVs should adapt to this dynamism and avoid ML models suited only for static networks.
\item \textbf{Directed:} CAVs must adhere to traffic regulations, which limit their mobility and network topology to the road infrastructure. Therefore, ML-based TMS methods for CAVs should account for whether CAV behaviors align with road topology constraints.

\item \textbf{Opportunistic:} IoVs frequently interact with unknown neighbors. Therefore, ML-based TMS for CAVs should be timely and include incentives to encourage greater node participation.

\item \textbf{Intermittent:} CAVs continuously form and disband networks. Therefore, ML-based TMS methods should accommodate this dynamic topology, for instance, by using dynamic GNNs that can integrate new nodes as they appear.

\item \textbf{Resource Availability:} Each CAV is equipped with limited computational resources. Therefore, edge and cloud computing are essential to support ML-based TMS methods that require substantial computational resources, and blockchain technology is needed for reliable data transmission.
\item \textbf{More motivations to build trust among vehicles}: While ~\textcolor{blue}{\cite{Hussain2022}} noted that vehicles have less motivation to build trust through membership due to automatic network formation within communication range, trust has become crucial in more scenarios like platoon leader selection, worker incentives in crowdsensing and choosing reliable edge nodes, etc.
\end{itemize}

\subsection{Layer1:Trust Data Layer Enhanced by ML for CAV}

i) \textbf{Trust-related data composition.} It refers to data structure and dimensions, including structured data like vehicle beacons, trajectories, sensors, detectors and maps, as well as unstructured data like radar, LiDAR, and images. Beacon data is widely used for its ease of acquisition and ability to reflect vehicle behavior. Multimodal data fusion is also gaining attention. ML methods excel at processing large-scale, diverse datasets.

Beacon data constitutes the fundamental broadcast information, featuring message types such as Basic Safety Messages (BSM), Map Data (MAP), Signal Phase and Timing (SPAT), Roadside Messages (RSM), and Personal Safety Messages (PSM). The standard format of BSM is defined in (\ref{eq:bsm}) defined in~\textcolor{blue}{~\cite{SAEJ2735}},~\textcolor{blue}{~\cite{ETSI2019}}.

\begin{multline}
\label{eq:bsm} 
\makebox[0pt][l]{\hspace*{\textwidth}\hspace*{\multlinegap}\hspace*{\displaywidth}(\ref{eq:bsm})}
{\rm{BSM}} = \{ {\rm{senderPseudo}}, SendTime, Pos, Spd, Acl, Hed, \\
PosNoise, SpdNoise, AclNoise, HedNoise, msgID\}
\end{multline}

where $senderPseudo$ denotes the pseudonym of sending CAVs, $Pos,Spd, Acl,Hed$ refers to the position, speed, acceleration, and heading of senders, respectively. Ditto for those terms with $Noise$.$SendTime$ and $msgID$ are the send time and message IDs of BSM.

ii) \textbf{Trust-related data generation and dissemination.} It refers to data sources and collection methods. In single autonomous vehicles, data is mainly generated by onboard sensors. In vehicle-road-cloud systems, data from multiple sources is transmitted to edge devices and cloud servers.

iii) \textbf{Trust-related data and trust value storage.} In centralized systems, a central server stores all trust data, while in distributed models, each device stores its own. Most CAV TMS prefer the distributed model, with distributed ML and blockchain enhancing privacy beyond traditional cryptography.

iv) \textbf{Trust-related data dissemination}. Before trust computation, ML methods must gather sufficient trust-related data. However, it can be hindered by selfish or malicious CAVs. Implementing ML in the incentive layer can mitigate this issue.

\begin{figure}
  \begin{center}
  \includegraphics[width=3.5in]{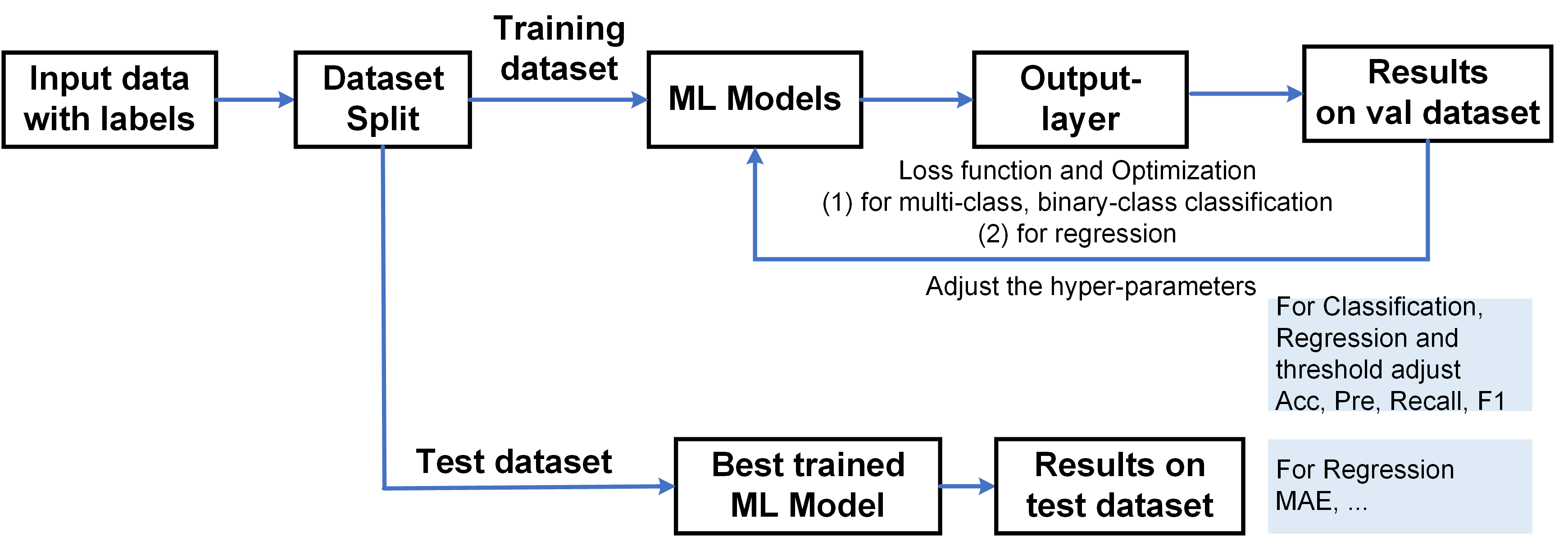}\\
  \caption{Basic process of ML for trust calculation}
  \label{fig:ml-tc}
  \end{center}
\end{figure}

\subsection{Layer2: Trust Computation Layer Enhanced by ML for CAV}
 
i) \textbf{Trust aggregation}. 
Trust-related data must be filtered from raw data to create standardized input. ML methods excel at analyzing complex, large-scale, and heterogeneous datasets.

ii) \textbf{Trust calculation}. Unlike rule-based or knowledge-based methods, ML approaches assess CAV trust through classification and regression by continuously analyzing and adapting to behavioral characteristics. It is described in ~\textcolor{blue}{Fig.\ref{fig:ml-tc}}.

iii) \textbf{Trust update}: 
ML methods like RL enable timely updates of trust values by continuously adapting to behavioral changes, allowing the trust model to evolve dynamically with time-triggered and event-triggered changes, ensuring accurate and relevant assessments.

iv) \textbf{Weight adjuster:}. 
Weights reflect the importance of various factors in trust calculations and must be adjusted based on user needs or environmental changes. ML algorithms can automatically learn and dynamically adjust these weights using user feedback or service results.

\subsection{Layer3: Trust Incentive Layer Enhanced by ML for CAV}
Incentives motivate nodes to ensure that they are willing to maintain the security and efficiency of the network while preventing them from having negative, speculative, or malicious motives.

i) \textbf{Incentives for negative, speculative motives}: Different from incentives for TMS that were classified into ex-ante incentive, ex-post incentive in~\textcolor{blue}{\cite{Wei2022}}, learning-based incentive methods are focused, such as DRL combined with traditional incentives, like auction game, contract, Stackelberg game, etc.  

iv) \textbf{Defend against trust-related attacks}: 
Trust-related attacks manipulate trust values to undermine TMS accuracy. For example, in a self-promotion attack (SPA), attackers inflate their trust values by falsely claiming positive behavior. In a whitewashing attack (WA), attackers obscure their negative past by displaying a series of positive behaviors. ML enhances it by improving anomaly detection capabilities, dynamically adjusting trust calculations, complex behavioral features abstraction and fusion of multi-source data.

\begin{table*}[h!]
\centering
\caption{Comparison of Non-ML-based TMS Methods and ML-based TMS Methods}
\begin{tabular}{|p{1cm}|p{1.5cm}|p{3.5cm}|p{4.5cm}|p{5cm}|}
\hline
\textbf{Objectives} & \textbf{Categories} & \textbf{Dis in Non-ML Methods} & \textbf{Ads in ML Methods} & \textbf{Challenges to ML-based TMS} \\ \hline
\multirow{5}{*}{\parbox{1cm}{Effectiveness}} & Collecting enough data & Statistic methods to cope with simple data source and simple data type. & Automatically learning patterns and trends from massive amounts of data & Variations in data quality among different sources and potential noise during data collection. Privacy. \\ \cline{2-5} 

 & \multirow{2}{*}{Accuracy for trust computation} & Bayesian inference-based TMS uses a priori assumptions or probability distributions. & Handles non-linear relationships and multi-dimensional data, mines complex patterns and behaviors, which significantly improves accuracy and predictive capability. & Poor interpretability; Results are sensitive to data bias and data imbalance. \\ \cline{3-4}
 &  & Fuzzy-based TMS & Adapt to data characteristics and environmental changes. & \\ \cline{2-5} 
 & Accuracy for supporting trust-related services & Only ML-based service for enhancing performance but absent trust methods for security and privacy protection. & Trust-related services keep the same data as ML-based services without extra data collection. ML-based TMS can enhance the system's robustness against potential attacks by real-time data processing and automated decisions. & Privacy, poor interpretability, continuous learning and updating, and computational resource demands. \\ \cline{2-5} 
 & Accuracy for incentive mechanisms on different conditions & Predefined models such as auction theory and game theory. & Allow them to deal with complex incentive problems to improve the overall efficiency and fairness of the system. & Affected by data fluctuations and model updates, which leads to instability of the strategies. Poor interpretability. It may take longer to learn and adapt to new behavioral patterns. It may amplify these unfair biases due to the bias of the dataset. \\ \hline
\multirow{4}{*}{Flexibility} & Flexibility to dynamic changes & Predefined models or rule-based systems that may not adapt quickly to changes. & Adapting to changes in network topology and environmental conditions & Requiring continuous learning and updating to handle dynamic changes. High computational costs are associated with frequent model retraining. \\ \cline{2-5} 
 & Different services with trust values at different granularities. & For different granularities, it should perform different methods. & Similar ML models can be adopted for different services, such as SVM and SVC. Adapt to changes in network topology and environmental conditions. & Hard to select the appropriate model for each service. Increased complexity in managing multiple models. \\ \hline
  \multirow{4}{*}{Reliability}
 &Stability for supporting trust-related services & Static models and predefined rules may not provide a stable service at different locations and environments.  & Dynamically adjust and maintain stability by continuously learning from new data.& High dependency on the quality and recency of the data used for training. Potential instability during model updates and retraining.  \\ \hline

\cline{1-5} 
\multirow{2}{*}{Efficiency}& Time complexity & May not meet real-time requirements & The architecture can be designed based on real-time requirements and arranged in different locations. &Using lightweighting and acceleration of ML. Ensuring real-time performance while maintaining accuracy. \\ \cline{2-5} 
 & Computing resource complexity & Only applied to small data volume. & Coping with large-scale data. Using model compression and distributed computing. & Having a high computational cost and may exceed on-board computing resource limits. \\ \cline{2-5} 
 & Storage resource complexity & Only applies to small data volumes. & Coping with large data volumes and more complex data types. Using data reduction techniques such as feature selection and dimensionality reduction. & Requiring large storage usage beyond onboard storage constraints. \\ \hline
\multirow{4}{*}{Security} & Defending cyber attack & Not adapt quickly to new threats. & Detecting and responding to cyber threats in real-time by learning from past attack patterns & Longer time-consuming. Ensuring the robustness of ML models against evolving threats. Performing adversarial attacks on ML models. \\ \cline{2-5} 
 & Defending trust-related attacks & Less effective against sophisticated attacks. & Identifying and mitigating trust-related attacks through continuous learning and pattern recognition & Model performance variation against attack variations. Poor interpretability. Performing adversarial attacks on ML models. \\ \cline{1-5}
 \multirow{4}{*}{Privacy}
 &Privacy-preserving & Traditional TMS methods do not provide adequate privacy protection and require current privacy methods. & ML methods like FL can enhance privacy through architectural redesign. & Balancing privacy with model accuracy and performance. Ensuring compliance with privacy regulations and standards. \\ \hline
\end{tabular}
\label{tab:MLwithTrad}
\end{table*}

\subsection{Objectives of ML-based TMS for CAVs}
A novel taxonomy for objectives in ML-based TMS for CAVs is introduced. This taxonomy, inspired by recent surveys on TMS~\textcolor{blue}{\cite{Liu2023BlockTMS}},~\textcolor{blue}{\cite{Wei2022}}, uniquely integrates the specific features of CAVs with the requirements of TMS and ML methods. Furthermore, it provides a comparative analysis of traditional non-ML TMS methods versus ML-based approaches, evaluated across six dimensions, as detailed in ~\textcolor{blue}{Table.\ref{tab:MLwithTrad}}.

\textbf{(1) Effectiveness of TMS:}
Effectiveness is the primary requirement for a sound TMS. It is judged from the following four aspects.

\begin{itemize}
    \item Enough data collections:
It ensures the system's ability to gather sufficient data to support the TMS effectively.
    
    \item Accuracy for trust computation:
The accuracy metric serves as a representative indicator for all other metrics. It ensures the accuracy of the trust calculation model, guaranteeing that trust values accurately reflect vehicle behavior.

   \item Accuracy for supporting trust-related services:
It ensures that the trust values produced by the TMS effectively support various trust-related services.

   \item Accuracy for incentive mechanisms on different conditions: 
It ensures that the TMS supports the implementation of incentives or penalties based on diverse behaviors, thereby fostering cooperative behavior.
\end{itemize}

Evaluation metrics for effectiveness are as follows:
\begin{itemize}
    \item Trust calculation regarded as a classification task:
\textit{a) Problem Formulations}: The classifier receives the state information of the vehicle as input and outputs a direct trust value indicating the trust value of the CAV. If this trust value is below a threshold, the CAV may be classified as malicious; if the trust value is above the threshold, the vehicle is considered trusted.
\textit{b) Evaluation Metrics}: Four basic metrics for classification are Accuracy, Precision, Recall and F1-score. Besides, more metrics are the Matthew Correlation Coefficient (MCC), Cohen's kappa, and the Receiver Operating Characteristic (ROC), etc.
\item Trust calculation regarded as a regression task:
\textit{a) Problem formulations}: It is regarded as a regression task to predict continuous trust values. 
\textit{b) Evaluation metrics}: Common metrics are Mean Squared Error (MSE), Root-MSE (RMSE), Mean Absolute Error (MAE), R-squared.
\end{itemize}

\textbf{(2) Flexibility of TMS: }
Given the mobility of CAV nodes and the dynamic operating environments, flexibility in TMS for CAVs is more critical than in static IoT systems. Flexibility serves as the foundation for other objectives and can be categorized into two key aspects.
\begin{itemize}
    \item Flexibility to dynamic changes: The TMS should adapt to changes in network topology, vehicle behavior, and environmental conditions with flexibility.
    \item Granularity of trust values for different services: It should provide fine-grained trust values (e.g., continuous values) for services that demand high precision, while coarse-grained trust values (e.g., binary or multi-class) for services that do not need detailed trust information.
\end{itemize}

Evaluation metrics for flexibility are as follows:
\begin{itemize}
    \item Adaptability metrics: a) Rate of model adjustment: It measures the speed at which the model adapts to changes (e.g., in network topology, vehicle behavior). b) Convergence speed: It indicates how quickly the model re-learns and stabilizes after a dynamic change.
    \item Scalability metrics: It assesses the model's ability to maintain performance as data types or volumes, node density, or attacker density increase.
    \item Continuous learning metrics: a) Frequency of model updates:  It evaluates how often the model is updated to reflect environmental changes. b) Online learning efficiency: It measures the effectiveness of continuous learning without requiring offline retraining.
\end{itemize}

\textbf{(3) Reliability of TMS: }
It is mainly in terms of its 
\begin{itemize}
    \item Ability to provide stable support:i.e., the TMS should maintain consistent and stable performance across various situations, providing accurate and reliable services without interruption.
    \item Interpretability of ML: As the ML theory evolves, the decision-making process of understanding the ML-based trust model can be enhanced. 
\end{itemize}

Reliability should be assessed using software reliability metrics such as availability, Mean Time to Recovery (MTTR), Mean Time Between Failures (MTBF), and failure rate. Interpretability metrics include decision boundary visualization, feature importance, Shapley value, etc. Besides, more metrics align with those used to evaluate effectiveness. 

\textbf{(4) Efficiency of TMS:}
It is crucial to ensure that TMS processes are responsive and resource-efficient. It encompasses several aspects:

\begin{itemize}
    \item Time complexity: The TMS should maintain low time complexity to support real-time decision-making. Higher time complexity is allowed for other service requirements.
    
    \item Computing resource complexity: 
    Some modules are designed to operate efficiently within the limited resources of onboard modules. For complex and long-term tasks, leveraging edge devices and optimizing architectural design is essential.

    \item Storage resource complexity: 
    The TMS should use efficient data storage structures to minimize storage needs while ensuring data accessibility and integrity.
\end{itemize}

While model performance evaluation is related to model accuracy, runtime performance evaluation determines the performance of the model when deployed in C-ITS. Models that consume a large amount of resources in detecting misbehavior may block resources used for normal services, leading to additional misbehavior.
Evaluation metrics are: 
\begin{itemize}
    \item ML Complexity:  It describes the relationship between the size of the data and the time using Big O notation.
    \item Communication and storage costs: It covers the network latency, and execution time, throughput, packet Delivery Ratio (PDR), memory usage, data storage capability, etc.
\end{itemize}

\textbf{(5) Security of TMS:}
\begin{itemize}
    \item Defending against cyber attacks: 
TMS can defend against various cyber attacks, such as DoS attacks, tampering attacks, and Sybil attacks. 
\item Defending trust-related attacks:
TMS should defend against trust-related attacks. 
\end{itemize}
Evaluation metrics are consistent with the effectiveness. 

\textbf{(6) Privacy preserving for TMS}
TMS collects large amounts of data, which covers identity privacy, location privacy and behavioral privacy, etc. Thus, privacy-preserving for TMS becomes critical 
under strict regulations, such as the General Data Protection Regulation (GDPR).
Common privacy-preserving measures are 
anonymization, perturbation, encryption (e.g., Homomorphic Encryption), differential privacy, Secure Multiparty Computation (SMC), Zero-Knowledge Proofs. Blockchain is not a method of privacy-preserving, but it provides a secure, transparent and hard-to-tamper-with way of storing and transmitting data that helps enhance privacy protection.FL is also hot research in privacy preservation, and extensions such as Federated DP~\textcolor{blue}{\cite{xu2023FDL}}, Federated SMC~\textcolor{blue}{\cite{chen2024FedSMC}} and Blockchain-based FL~\textcolor{blue}{\cite{sameera2024FLBlock}} have also been proposed. In addition, TMS needs to consider pseudonymization mechanisms for vehicles. Evaluation metrics are consistent with the effectiveness. Some methods have their own quantitative privacy-preserving metrics, such as the privacy budget in Federated DP.

\section{Applications Scenarios of TMS for CAVs}
Based on the analytical framework established in Section IV, this survey systematically examines the implementation of ML-based TMS across five CAV application scenarios, which is significant for more practical ITS studies.
These scenarios are categorized according to the four Cooperative Driving Automation (CDA) classifications defined in SAE J3216  (status-sharing, intent-sharing, agree-seeking, and prescriptive). It is supplemented by a dedicated section on novel TMS advancements.
Due to space limitations, results for certain categories are presented textually and visually through graphs and tables, while others are described textually.

The investigation pursues three objectives derived from Section IV's framework and metrics. (a) Through comprehensive literature documentation within each application scenario, we identify the specific techniques and analyze emerging research trends; (b) Tabular comparative analysis facilitates systematic cross-study evaluation of methodological approaches;
(c) Critical assessment demonstrates how ML enhances traditional TMS approaches, while simultaneously mapping the evolutionary trajectory of this research domain. To evaluate the current studies, the metrics are against six dimensions from Section IV.E.

a) Effectiveness: 
A study is rated as \textbf{highly effective} if it evaluates the accuracy of TMS, employs data fusion techniques to enhance TMS performance, and demonstrates improvements through TMS.

b) Flexibility: A study is rated as \textbf{highly flexible} if it investigates dynamic trust update mechanisms and proposes incentive designs to improve system adaptability.

c) Reliability: A study is rated as highly reliable if it analyzes the impact of trust fluctuations on service quality and develops countermeasures against trust-related attacks.

d) Efficiency: A study is rated as highly efficient if it measures the computational and time costs of TMS operations.

e) Security: A study is rated as highly secure if it identifies and analyzes at least four distinct types of trust-related attacks in TMS.

f) Security: A study is rated as highly private if it focuses on privacy-preserving techniques within TMS.

\subsection{TMS in Data Dissemination}
\textbf{\textit{Scenario description:}} Data dissemination, also known as content distribution or data sharing, is essential in IoV for reducing accident risks and improving traffic management efficiency. It can be categorized as follows: a) Motivations: opportunistic, vehicle-assisted, and cooperative data dissemination; b) Event types: event or on-demand dissemination, periodic dissemination, and hybrid dissemination; c) Range and distance: multi-hop or single-hop dissemination; d) Applications: delay-sensitive safety applications and delay-tolerant comfort and infotainment. The last classification criterion was chosen in our survey. Various TMS methods have been proposed to address their different features.

\begin{table*}[htbp]
\centering
\caption{Comparison of TMS methods for delay-sensitive safety data dissemination in CAV}
\renewcommand{\arraystretch}{1}
\begin{tabular}{p{0.3cm} p{0.3cm} p{1.2cm} p{5cm} p{3cm} p{0.5cm} p{0.5cm} p{0.5cm} p{0.3cm} p{0.3cm} p{0.3cm} p{0.3cm}}
\toprule
\multicolumn{6}{c}{} & \multicolumn{4}{c}{\textbf{Metrics for TMS}} \\
\cmidrule(lr){7-12}
\textbf{Ref.} & \textbf{Time} & \textbf{Data Layer} & \textbf{Computation Layer} & \textbf{Incentive Layer} & \textbf{Data} & \textbf{Effect} & \textbf{Fle} & \textbf{Rel} & \textbf{Eff}  & \textbf{Sec} & \textbf{Pri}\\
\midrule
%1
\colorbox[HTML]{CCFFCC}{\textcolor{blue}{\cite{Wang2024PS}}}
& 2024 & Distributed & Bayesian interference for direct trust & Stackelberg Equilibrium for ex ante incentives & Matlab & \LEFTcircle & High & \LEFTcircle & × & × & × \\
%2
\colorbox[HTML]{CCFFCC}{
  \textcolor{blue}{\cite{qi2024hybrid}}
}
 & 2024 & Distributed & Statistical weighted method for direct trust and trust certificate & None & Veins & High & \LEFTcircle & High & × & \LEFTcircle & × \\

%3
\colorbox[HTML]{CCFFCC}{
  \textcolor{blue}{\cite{Li2023Ann}}
}
 & 2023 & Blockchain & Not mentioned the specific method but mentioned a sum of  vehicle's resource contribution and reputation value & Bidding Mechanism, Deposit System & Veins & \LEFTcircle & High & \LEFTcircle & High &× & \LEFTcircle  \\

\colorbox[HTML]{CCFFCC}{\textcolor{blue}{\cite{Yang2023Block}}}
 & 2023 & Blockchain & Dirichlet-based probabilistic model & Setting penalties but not incentives & Taxi GPS & \LEFTcircle & \LEFTcircle & \LEFTcircle & High & \LEFTcircle &\LEFTcircle\\

\colorbox[HTML]{CCFFCC}{\textcolor{blue}{\cite{Yuan2023}}} & 2023 & Blockchain & Statistical weighted method for direct trust and indirect trust & Lower the trust value for lazy vehicles & Geth & \LEFTcircle & High & High & High & \LEFTcircle &\LEFTcircle\\

\colorbox[HTML]{CCFFCC}{\textcolor{blue}{\cite{Fan2023}}} & 2023 & Blockchain & Statistical weighted method for direct trust and indirect trust & Pos and PBFT-based incentives & Veins & \LEFTcircle & \LEFTcircle & High & High & High &\LEFTcircle\\

\colorbox[HTML]{CCFFCC}{\textcolor{blue}{\cite{Shen2024Block}}} & 2024 & Blockchain & Statistical weighted method for direct trust, indirect trust, RSU recommend trust & None & Veins & \LEFTcircle & \LEFTcircle & \LEFTcircle & High & × & High\\

\cmidrule{1-12}
\colorbox[HTML]{FFCC99}{\textcolor{blue}{\cite{Khatri2023}}} & 2023 & Blockchain & \textbf{Clustering algorithm} & None & Pygame & \LEFTcircle & \LEFTcircle & \LEFTcircle & \LEFTcircle & × &\LEFTcircle\\

\colorbox[HTML]{FFCC99}{\textcolor{blue}{\cite{Zhang2024Bay}}} & 2024 & Blockchain & \textbf{GaussianNB classifier} & None & Python & \LEFTcircle & \LEFTcircle & High & High & High &\LEFTcircle\\

\colorbox[HTML]{FFCC99}{\textcolor{blue}{\cite{Xia2023Dis}}} & 2023 & Distributed & Statistical weighted method but weight adjusted by \textbf{RL} & \textbf{RL-based incentive(but only Q-Learning)} & SUMO & High & \LEFTcircle & \LEFTcircle & High & \LEFTcircle &\LEFTcircle\\

\colorbox[HTML]{FFCC99}{\textcolor{blue}{\cite{Sarker2023Dis}}} & 2023 & Distributed & Bayes-based direct trust and Yager's rule-based indirect trust & \textbf{RL-based incentive (but only Q-Learning)} & ns-2 & \LEFTcircle & \LEFTcircle & \LEFTcircle & High & High & ×\\
\colorbox[HTML]{FFCC99}{\textcolor{blue}{\cite{Tao2024DQN}}} & 2024 & Distributed & Statistical weighted method and trust as factor for path prediction & None & Python & \LEFTcircle & \LEFTcircle & \LEFTcircle & High & × &×\\

\colorbox[HTML]{FFCC99}{\textcolor{blue}{\cite{wang2023Fed}}}  & 2023 & Distributed & Beta-Bayesian probabilistic model \textbf{and FL} & None & Matlab & \LEFTcircle & \LEFTcircle & \LEFTcircle & High & × &\LEFTcircle\\

\bottomrule[1.5pt]
\label{tab:dissemination}
\end{tabular}
\begin{minipage}{\textwidth}
\footnotesize
\raggedright
\textcolor{black}
{Notes:  \LEFTcircle: Semi-satisfied. ×: Not mentioned. 
Colors: 
\colorbox[HTML]{CCFFCC}{[X]:Non-ML-based TMS for CAV};
\colorbox[HTML]{E0E0E0}{[X]:Non-ML-based TMS for other fields}; 
\colorbox[HTML] {FFCC99}{[X]:ML-based TMS for CAV}; 
\colorbox[HTML] {C0C0C0}{[X]:ML-based TMS for Other Fields}};
\end{minipage}
\vspace{-8pt}
\end{table*}

\subsubsection{\textbf{TMS in delay-sensitive safety applications}}
These applications require rapid and reliable data transmission to ensure the safety of vehicles and passengers. Examples include collaborative autonomous driving, accident event dissemination, unsignalized intersection control, messages from special vehicles like ambulances and police vehicles, etc. 

Essential criteria for TMS implementation.
These applications are usually disseminated periodically, enabling CAVs not to be required to the specific location~\textcolor{blue}{\cite{shah2022datadis}}. Besides, its effectiveness relies on the security provided by TMS. This section focuses on those works of trust in data dissemination and other sections will investigate those considering performance enhancements. The comparison is summarized in~\textcolor{blue}{Table.\ref{tab:dissemination}}. 

(1) \textit{Lessons from traditional TMS methods. }
a) \textit{TMS can be a hybrid data trust and entity trust.} An entity-centric model by~\textcolor{blue}{\cite{Wang2024PS}} was proposed for collaborative autonomous driving. A hybrid trust model was proposed for emergency message dissemination~\textcolor{blue}{\cite{qi2024hybrid}}. b) \textit{Recent advances have seen a growing interest in blockchain-based TMS methods for IoV.} Main studies are as follows: blockchain-based TMS for resource integration and incentives~\textcolor{blue}{\cite{Li2023Ann}}, and blockchain TMS framework~\textcolor{blue}{\cite{Yang2023Block}}, consortium blockchain-based TMS~\textcolor{blue}{\cite{Fan2023}}. c) \textit{Some solutions were on enhancing the time efficiency of TMS.} Yuan et al.,~\textcolor{blue}{\cite{Yuan2023}} supported TMS by reducing network congestion and avoiding the propagation of duplicate data. d) \textit{Some solutions were on data privacy.} Shen et al.,~\textcolor{blue}{\cite{Shen2024Block}} incorporated blockchain-based multi-party trust evaluations for data privacy preservation.

(2) \textit{Recent ML-based TMS methods}. These methods can be categorized as follows.
\begin{itemize}
    \item \textbf{Clustering:} The clustering algorithm itself does not directly generate trust values but indirectly helps in evaluating and calculating trust values by grouping entities with similar characteristics. VEMCA~\textcolor{blue}{\cite{Khatri2023}} was a clustering-based algorithm for IoV event messages, where the message trust was the ratio of true messages to total messages within a given time after clustering. 
    \item \textbf{Classifier:} The categorization of CAVs' behaviors enables an effective assessment of their trust. The GaussianNB classifier was used to determine vehicle direct trust values ~\textcolor{blue}{\cite{Zhang2024Bay}}, and indirect trust was not considered.
    \item \textbf{RL:}
The applications of RL in data dissemination are weight adjustment for trust updating; trust as RL's state and RL indirectly impacts; TMS for some decision-making problems ~\textcolor{blue}{\cite{Tao2024DQN}},{\cite{Xia2023Dis}},\textcolor{blue}{\cite{Liu2024EffMsg}}; and play role in incentive mechanisms{\cite{Xia2023Dis}},~\textcolor{blue}{\cite{Sarker2023Dis}}.
    \item \textbf{FL: } FL offers distributed training for TMS. Wang et al.~\textcolor{blue}{\cite{wang2023Fed}} proposed a personalized FL-based node TMS for CAVs on statistical trust calculation.
\end{itemize}

Advances and limitations of prior work are summarized as follows. a) Attack defense: While the trust values were optimized, VEMCA cannot resist DoS attacks~\textcolor{blue}{\cite{Khatri2023}}. b) Service improvement: Trust-based information dissemination policies were discussed in traffic guidance scenarios and accident warnings scenarios~\textcolor{blue}{\cite{Xia2023Dis}}.
c) From using traditional ML to optimized ML. 
The RL method was traditional Q-Learning and was not compared to many RL methods, where the DESQN method~\textcolor{blue}{\cite{Tao2024DQN}} and Cooperative Hierarchical Attentional RL (CHA) improved this. However, the trust evidence was still based on statistical approaches, though the trust-sharing process and path prediction were based on the RL method. Liu et al.,\textcolor{blue}{\cite{Liu2024EffMsg}} proposed a CHA framework for hybrid cooperative driving based on V2V, including calculating factors against jamming attacks.

\subsubsection{\textbf{TMS for comfort and infotainment applications.}}
\textit{Scenario description:} These applications include personalized Points of Interest (POI) recommendations, music, news, and more. TMS enhances the effectiveness of these services by preventing misinformation and improving content filtering. 

\textit{Essential requirements for TMS implementation.}
Although TMS in these applications has less strict time constraints like delay-sensitive safety applications, they deal with various unstructured data types, such as text, audio, video, and user interaction data. User privacy protection is crucial~\textcolor{blue}{\cite{Wang2019sIoVPriSur}}. Besides, trust in these applications is often associated with social IoV, enabling drivers and passengers to socially interact with other vehicles sharing similar interests, backgrounds, or destinations. It is common among taxis or private cars that follow routine routes daily.

(1) \textit{Lessons from traditional TMS methods. }
Traditional TMS solutions have been proposed for social IoV, incorporating social trust besides direct trust and indirect trust, such as subjective logic~\textcolor{blue}{~\cite{Cheng2019SubLog}}, fuzzy logic ~\textcolor{blue}{\cite{Zhao2022EntitySocial}}, and statistically weighted methods~\textcolor{blue}{\cite{Li2020Incentive}}, ~\textcolor{blue}{\cite{Li2024VideoT}},~\textcolor{blue}{\cite{Mada2019VideoT}}. Besides the classification methods, the trust-related data is also different. TEAD~\textcolor{blue}{\cite{Li2024VideoT}} was designed for video sharing D2D networks, using the correctness of video content as trust evidence. Similarly, Mada et al.,~\textcolor{blue}{\cite{Mada2019VideoT}} proposed a trust-based framework for video content delivery.
TEAD~\textcolor{blue}{\cite{Li2024VideoT}} used D2D data while the work in~\textcolor{blue}{\cite{Mada2019VideoT}} used video data; neither was specifically designed for social IoV but for the general IoT.

(2) \textit{Recent ML-based TMS methods}. 
Fewer works in social IoV currently adopt ML-based TMS features. Although datasets and spatial-temporal features in social IoV are different, insights can be drawn from research in social networks and social IoT. These methods can be categorized as follows:
\begin{itemize}
    \item \textbf{ Clustering:} Jay et al.~\textcolor{blue}{\cite{Jay2019MLTMSsocial}} proposed a K-means clustering algorithm to visualize trust boundaries in the feature space, classifying data points with trust attributes into clusters labeled as "trust" or "untrust."

    \item \textbf{Classifier}: C-DeepTrust integrated feature abstraction using MLP and LSTM networks into a composite feature vector~\textcolor{blue}{\cite{Wang2023socialclass}}. Cosine similarity then converted these features into trust probabilities via the softmax function, with higher cosine similarity corresponding to higher trust values based on social homogeneity theory.
    \item \textbf{Regression}: Deng et al.~\textcolor{blue}{\cite{Deng2017SocialDL}} proposed a trust-aware recommendation approach, including pretraining with the deep autoencoder, social trust ensemble, and regularization with community effect.
    \item \textbf{GNN}: Several GNN-based methods have been proposed, including Guardian~\textcolor{blue}{\cite{lin2020guardian}}, GATrust~\textcolor{blue}{\cite{Jiang2023GATrust}}, TrustGNN~\textcolor{blue}{\cite{Huo2023TrustGNN}}, GainTrust~\textcolor{blue}{\cite{xu2023attsocial}}, and TrustGuard~\textcolor{blue}{\cite{Wang2024TrustGuard}}. For instance, compared to Guardian and TrustGNN, GainTrust and TrustGuard optimized GNN for dynamic social networks. TrustGuard also defended against trust-related attacks and added explainability, while GainTrust did not consider them.
    
\end{itemize}

\subsection{TMS in Routing}
\textbf{\textit{Scenario description:}} IoV routing refers to the process and methodology by which packets are transmitted from a source to a destination (another vehicle, infrastructure, or remote server). In our survey, the source is only considered to be vehicles. IoV routing can be categorized into route discovery, path selection, and route maintenance. These routing are to determine which path to choose for transmitting data. Common criteria for routing include shortest path, minimum number of hops, maximum bandwidth, and minimum latency. Multi-hop task routing based on MADDPG~\textcolor{blue}{\cite{Deng2024Route}} was proposed under the vehicle-assisted collaborative edge computing framework, which allows tasks to be efficiently routed and offloaded through multiple intermediate nodes for optimal resource allocation and efficient task completion. But the security problem was not considered. Thus, ensuring that efficiency and performance are met while also providing the necessary security, such as trust management, selects the path with the highest level of trust. The comparison of this part is summarized in~\textcolor{blue}{Table.\ref{tab:route}}. 

\subsubsection{Securing and Optimizing Route Protocol and Route Process}
\textit{(1) Lessons from traditional TMS methods.}
Main studies are as follows: statistically weighted methods~\textcolor{blue}{\cite{sho2023tgrv}},
subjective logic~\textcolor{blue}{\cite{Mirzadeh2023MalRoute}},
and graph-based methods~\textcolor{blue}{\cite{Vegni2024Social}}. How trust metrics should be involved in the routing process can be learned.
Specifically, TGRV was proposed in~\textcolor{blue}{\cite{sho2023tgrv}} to select trustworthy next-hop nodes and restrict malicious vehicles from participating in the routing process. 
In~\textcolor{blue}{\cite{Mirzadeh2023MalRoute}}, the trust metrics were freshness, reliability and routing path length of the packet and the path with a high trust level was selected for forwarding the packet. 
In~\textcolor{blue}{\cite{Vegni2024Social}}, the cluster heads analyzed social characteristics like nodal degree, centrality degree, and friendship degree and then selected the most trusted node as the next hop-forwarder.
Besides, a trust-based routing protocol for VANET can be assisted by RSU, such as RTRV~\textcolor{blue}{\cite{azizi2024rtrv}}.

(2) \textit{Recent ML-based TMS methods}.
Although some studies were for underwater IoT and underwater acoustic sensor networks (UASNs), the nodes also have similar features like dynamic topology, but also have differences. For example, some measures for meeting lightweight capability may not be in CAV.

\begin{itemize}
    \item \textbf{Clustering:}
Fan et al.,~\textcolor{blue}{\cite{fan2019trust}} proposed a routing method using a trust model and game-based incentives. The trust model combines direct and indirect trust through an uncertain deductive method, and an attribute-weighted K-means clustering method is used to identify legitimate messages.
    \item \textbf{Classifier:}
A light-weight trust-aware multicast routing protocol (LTMRP) was proposed in~\textcolor{blue}{\cite{Xia2019Route}}, where subjective trust is based on SCGM(1,1) Weighted Markov Prediction and recommendation trust assessment method based on feedback mechanism. And it can establish secure and reliable communication paths by selecting trusted relay vehicles. In other fields, Zhu et al.,~\textcolor{blue}{\cite{Zhu2024underwaterRoute}} proposed a routing protocol based on GMM-HMM-LSTM for the underwater IoT. 

   \item \textbf{RL methods:}
   A DQN-based TMS for software-defined VANETs was proposed in ~\textcolor{blue}{\cite{Zhang2022VRoute}} for getting the optimal selection policy of reliable communication links.
   Li et al.,~\textcolor{blue}{\cite{Li2024underwater}} proposed a Q-Learning-Assisted trust routing scheme for SDN-Based UASNs.
   
\end{itemize}

\subsubsection{Cluster Head Selection}
Cluster results in the distributed formation of hierarchical network structures by grouping vehicles together based on correlated spatial distribution and relative velocity~\textcolor{blue}{\cite{Cooper2017Cluster}}. Cluster methods can be used to improve routing scalability in IoV. Cluster headers can act as a collection and distribution center for routing information, helping to determine the best path for data transmission. Trust is critical criterion for cluster head selection.

(1) \textit{Lessons from traditional TMS methods}.
Currently, several TMS methods for clustering have used statistically weighted methods by direct trust and indirect trust~\textcolor{blue}{\cite{kadam2023CH}}, by Knowledge-Experience-Reputation{\cite{Awan2020}}, the interval type-2 fuzzy logic~\textcolor{blue}{\cite{Yang2023GAN}}. 

(2) \textit{Recent ML-based TMS methods.}
ML-based TMS methods for cluster head selection are few in CAV but can be inspired by industrial WSNs and UASNs.
\begin{itemize}
    \item \textbf{GAN:}
A GAN-based trust redemption model for industrial WSN was constructed~\textcolor{blue}{\cite{Yang2023GAN}} to avoid the permanent isolation of normal nodes due to false alarms. 
   \item \textbf{Classifier:} Han et al.,~\textcolor{blue}{\cite{Han2019Water}} proposed an SVM-based trust prediction model for UASNs.
\end{itemize}

\subsubsection{Relay Vehicle Selection}
Relay vehicles can improve the coverage of communications, increase the redundancy of signals, and increase the rate of data transmission, etc. TMS helps choose reliable relay vehicles that are exposed to varying threats. The trust values were based on performance metrics such as the bit error rate (BER) in ~\textcolor{blue}{\cite{Ghourab2022}}.

\begin{table*}[htbp]
\centering
\caption{Comparison of TMS methods in routing}
\renewcommand{\arraystretch}{1}
\begin{tabular}{p{0.5cm} p{0.5cm} p{1.2cm} p{5cm} p{2cm} p{1cm} p{0.5cm} p{0.5cm} p{0.3cm} p{0.3cm} p{0.3cm} p{0.3cm}}
\toprule
\multicolumn{6}{c}{} & \multicolumn{4}{c}{\textbf{Metrics for TMS}} \\
\cmidrule(lr){7-12}
\textbf{Ref.} & \textbf{Time} & \textbf{Data Layer} & \textbf{Computation Layer} & \textbf{Incentive Layer} & \textbf{Data} & \textbf{Effect} & \textbf{Fle} & \textbf{Rel} & \textbf{Eff}  & \textbf{Sec} & \textbf{Pri}\\
\midrule

%32 TGRV: A trust-based geographic routing protocol for VANETs
\colorbox[HTML]{CCFFCC}{\textcolor{blue}{\cite{sho2023tgrv}}} & 2023 & Distributed & Combination trust by direct trust and recommendation trust with trust decay & None & Veins & \LEFTcircle & \LEFTcircle & \LEFTcircle & High & High & × \\

%41 Filtering Malicious Messages by Trust-Aware Cognitive Routing in Vehicular Ad Hoc Networks
\colorbox[HTML]{CCFFCC}{~\textcolor{blue}{\cite{Mirzadeh2023MalRoute}}} & 2023 & Blockchain & Distributed	subjective logic and some trust factors & None & VSimRTI & \LEFTcircle & \LEFTcircle & High & × & × & ×\\

%93 A Reputation-Based Trustworthiness Concept for Wireless Networking in Vehicular Social Networks
\colorbox[HTML]{CCFFCC}{~\textcolor{blue}{\cite{Vegni2024Social}}} & 2024 & Blockchain & Hybrid Reputation metrics, probability of successful transmission& None & MatLab & \LEFTcircle & \LEFTcircle & \LEFTcircle & × & × & ×\\

%94 RTRV: An RSU-assisted trust-based routing protocol for VANETs
\colorbox[HTML]{CCFFCC}{~\textcolor{blue}{\cite{azizi2024rtrv}}} & 2024 & Hybrid & Direct trust by vehicle, indirect trust by RSU and TA without trust decay & Penalty factor & Veins & \LEFTcircle & \LEFTcircle & High & High & High & \LEFTcircle \\

%95 On trust models for communication security in vehicular ad-hoc networks
\colorbox[HTML]{FFCC99}{\textcolor{blue}{\cite{fan2019trust}}} & 2019 &  Distributed &	Certain-Factor for direct trust, Fuzzy C-mean clustering for indirect trust & Dynamic game model & MobiSim, NS-2 & \LEFTcircle & High & High & × & × & ×\\

%96 An Attack-Resistant Trust Inference Model for Securing Routing in Vehicular Ad Hoc Networks
\colorbox[HTML]{FFCC99}{~\textcolor{blue}{\cite{Xia2019Route}}} & 2019 & Hybrid & Direct trust, indirect trust,role-based trust, global trust, reliability assessment & Penalty factor & SUMO & High & \LEFTcircle & High & × & High & ×\\

%97 An Efficient Secure and Adaptive Routing Protocol Based on GMM-HMM-LSTM for Internet of Underwater Things
\colorbox[HTML]{C0C0C0}{~\textcolor{blue}{\cite{Zhu2024underwaterRoute}}} & 2024 & Blockchain & Distributed	GMM-HMM, LSTM; Energy trust, communication trust and node trust for underwater IoT & None & None & \LEFTcircle & \LEFTcircle & High & × &×&\LEFTcircle\\

%98 Software-Defined Vehicular Networks With Trust Management: A Deep Reinforcement Learning Approach
\colorbox[HTML]{FFCC99}{~\textcolor{blue}{\cite{Zhang2022VRoute}}} & 2022 & Centralized by SDN & Statistically weighted method for direct trust of the control packets, the data packets. Trust is the state of DQN. & None & MATLAB & \LEFTcircle & High & High & High &×&×\\

%99 SDN-QLTR: Q-Learning-Assisted Trust Routing Scheme for SDN-Based Underwater Acoustic Sensor Networks
\colorbox[HTML]{C0C0C0}{~\textcolor{blue}{\cite{Li2024underwater}}}&	2024&Centralized by SDN&	Link trust,data trust,node trust for UASNs&	None& MATLAB &\LEFTcircle &\LEFTcircle &High	&×& ×& × \\

\cmidrule{1-12}
%33 Cybersecurity threats mitigation in Internet of Vehicles communication system using reliable clustering and routing
\colorbox[HTML]{CCFFCC}{~\textcolor{blue}{\cite{kadam2023CH}}}&	2023&	Distributed	&Trust calculated by direct trust, indirect trust and as adaptive function in ACO algorithms	& None	& NS2 &	\LEFTcircle & \LEFTcircle & \LEFTcircle	& \LEFTcircle &×& \LEFTcircle\\

%101 StabTrust—A Stable and Centralized Trust-Based Clustering Mechanism for IoT Enabled Vehicular Ad-Hoc Networks
\colorbox[HTML]{CCFFCC}{{\cite{Awan2020}}}&	2020&	Centralized&	K-R-E&	None & NS2&\LEFTcircle & \LEFTcircle & \LEFTcircle	& \LEFTcircle&\LEFTcircle&×\\	

%39 Generative Adversarial Learning for Trusted and Secure Clustering in Industrial Wireless Sensor Networks
\colorbox[HTML]{C0C0C0}{~\textcolor{blue}{\cite{Yang2023GAN}}}&	2023&	Centralized&	Fuzzy logic and GAN	&None&	None & \LEFTcircle & \LEFTcircle & \LEFTcircle & \LEFTcircle&×&×\\

%102 A Synergetic Trust Model Based on SVM in Underwater Acoustic Sensor Networks
\colorbox[HTML]{C0C0C0}{\textcolor{blue}{\cite{Han2019Water}}} &	2019&	Distributed&	communication trust, packet trust and energy trust by KNN and SVM&	None&	MATLAB&	Middle&	Middle&	High&	×&\LEFTcircle&×\\
\cmidrule{1-12}
%103 Blockchain-Guided Dynamic Best-Relay Selection for Trustworthy Vehicular Communication
\colorbox[HTML]{C0C0C0}{~\textcolor{blue}{\cite{Ghourab2022}}}&	2022&	Blockchain&	Statistically weighted method for best relay selection	& By blockchain &	MATLAB&	\LEFTcircle&	\LEFTcircle&	High&	×&	×&	\LEFTcircle\\
\bottomrule[1.5pt]
\label{tab:route}
\end{tabular}
\begin{minipage}{\textwidth}
\footnotesize
\raggedright
\textcolor{black}
{Notes:  \LEFTcircle: Semi-satisfied. ×: Not mentioned.  
Colors: 
\colorbox[HTML]{CCFFCC}{[X]:Non-ML-based TMS for CAV};
\colorbox[HTML]{E0E0E0}{[X]:Non-ML-based TMS for other fields}; 
\colorbox[HTML] {FFCC99}{[X]:ML-based TMS for CAV}; 
\colorbox[HTML] {C0C0C0}{[X]:ML-based TMS for Other Fields}}
\\
\end{minipage}
\vspace{-8pt}
\end{table*}

\subsection{TMS in Vehicular Edge Computing}
These methods are discussed by edge offloading entity selection, VNE-based IoV and SDN-based IoV.

\subsubsection{Edge Offloading Entity Selection}

\textbf{\textit{Scenario description:}}
Vehicular edge computing (VEC) is categorized into fog computing, MEC computing, and micro cloud computing. The equipment of VEC includes edge nodes and edge servers. Trust is a key factor in the edge-offloading decision-making process and helps to optimize resource allocation to ensure that computational tasks are offloaded to well-performing and reliable edge equipment, providing a better user experience.

(1) \textit{Lessons from traditional TMS methods}.
Traditional TMS methods have been studied in edge computing by the Bayesian inference-based method~\textcolor{blue}{\cite{qure2020trustedge}}, a multi-feedback trust aggregation model{\cite{Kong2022Edge}}, statistically weighted average~\textcolor{blue}{\cite{mao2023edge}}, dual fuzzy logic{\cite{ali2024edge}} and subjective logic ~\textcolor{blue}{\cite{boun2024block}}. Specifically, QoS trust and the social trust were aggregated into global trust for the trust among RSU-CAV and RSU-RSU for securing task offloading~\textcolor{blue}{\cite{mao2023edge}}, where the weight parameters were set manually. 
Ali et al., {\cite{ali2024edge}} proposed a zero-trust management model for MEC and trust-aware task offloading. 
A trust assessment based on task execution feedback was proposed, adjusting the trust score with an incomplete Beta function, where a hedonic coalition game allowed vehicles to form cooperative groups for task offloading and execution, considering trust score, computational ability, and route matching {\cite{Pratap2024Edge}}. 
Wang et al.,{\cite{wang2025edge}} proposed a two-tier trust evaluation method to obtain accurate trust values for computation offloading.

(2) \textit{Recent ML-based TMS methods}.
ML-based TMS methods have been used for CAV.

\begin{itemize}
    \item \textbf{Classifier:} In~\textcolor{blue}{\cite{Mao2021Cloud}}, predicting the trust rate of a given cloud service is regarded as evaluating its QoS attributes and history, where several Back-Propagation Neural Networks (BPNNs) are used to form an integrated model, and PSO is used to obtain the optimal aggregation weights. The problem of trust prediction of IoT service consumers to edge computing nodes is modeled as a distributed optimization problem in {\cite{Abey2023edge}}, i.e., Network Lasso problem and a distributed trust prediction model is proposed based on the Stochastic Alternating Direction Method of Multipliers(S-ADMM).
    \item \textbf{RL:} In {\cite{Ren2021BlockRL}}, the A3C DRL algorithm was utilized to dynamically make optimal offloading and migration decisions by combining trust calculation and resource matching of vehicle nodes.
    
    \item \textbf{FL methods:}
The FEDQ-Trust method {\cite{bai2024fedq}} was proposed to select a subset of IoT MEC environments participating in federated learning by combining DQN on top of FedEM and using the trained global model to predict the trust value of IoT service instances. 
\end{itemize}

\subsubsection{Trust in VNE-based IoV}

\textbf{\textit{Scenario description:}}Virtual Network Embedding (VNE) is not part of edge computing but edge computing-enabling technology. 
VNE is a network virtualization technique that allows multiple virtual networks, with virtual nodes and links, to share a physical network infrastructure called the substrate network, which consists of substrate nodes and links. VNE can flexibly allocate resources to different vehicles, create virtual network slices to meet specific QoS requirements, support multiple service providers and users, and dynamically remap virtual nodes and links to achieve fault recovery and fault tolerance. In addition to considering constraints like resource limitations, QoS, and Quality of Information (QoI) for IoV such as~\textcolor{blue}{\cite{Fan2021VNE}}, VNE algorithms also take trust values into account, such as~\textcolor{blue}{\cite{Zhang2024VNE}}.

(1) \textit{Lessons from traditional TMS methods}.
Some academics have studied trust in VNE, where identifying and selecting trust nodes can defend against potential attacks, reduce conflicts and interference, and improve overall network resource utilization efficiency. Certain researchers have discussed the trust-aware VNE for industrial IoT~\textcolor{blue}{\cite{Zhang2020VNE}},~\textcolor{blue}{\cite{Rezae2024VNE}}. 
The trust calculation method of~\textcolor{blue}{\cite{Zhang2020VNE}} was information entropy TOPSIS, and the trust evaluation subject was substrate nodes, which aimed to ensure that virtual nodes can only be embedded on substrate nodes that meet their security requirements. Nevertheless, the authors \textcolor{orange}{did not} examine any trust or security level in the link mapping phase, making their approach a partially trust-aware algorithm claimed by~\textcolor{blue}{\cite{Rezae2024VNE}}.
Rezae et al.,~\textcolor{blue}{\cite{Rezae2024VNE}} considered both the trust levels of virtual nodes and substrate nodes, achieving efficient resource allocation and secure network performance optimization. 

(2) \textit{Recent ML-based TMS methods.}.
ML-based TMS has begun to be used for VNE in CAV. 
\begin{itemize}
    \item \textbf{RL. }Zhang et al.,~\textcolor{blue}{\cite{Zhang2024VNE}} evaluated the trust of substrate nodes using static security and activity levels, where the trust calculation was not based on ML but the DRL decisions can indirectly influence the trust calculation results by adjusting the weights of these trust parameters, which enables VNE resource allocation decisions that consider both performance and security.
\end{itemize}

\subsubsection{Trust in SDN-based IoV}
\textbf{\textit{Scenario description:}}SDN is also edge computing-enabling technology. SDN separates the control plane from the data plane, where centralized control and programmability provide the basis for VNE implementation. The flexibility and programmability provided by SDN allow networks in edge computing environments to quickly adapt to changes, such as dynamically adjusting network configurations to meet the demands of edge services. The importance of SDN security has been stressed in the future direction of a survey of ML applied in SDN in 2019~\textcolor{blue}{\cite{Xie2019SDNSurvey}}, but trust management was not covered. Additionally, SDN applications in IoV and SDN integrated with edge computing in IoV have improved a lot in recent years, such as SDN-based emergency social IoV~\textcolor{blue}{\cite{Sachan2024SDN}}, resource management by SDN integrated with fog computing~\textcolor{blue}{\cite{Nahar2024SDN}},~\textcolor{blue}{\cite{Liu2024SDN}}, etc. However, security was not considered in these methods. For instance,  Deng et al.,~\textcolor{blue}{\cite{Deng2024SDN}} summarized the types of topological attacks in SDN/SDVN, developing new defense mechanisms as future work.

Sub-scenario 1. \textbf{TMS enhanced by SDN.} 
It can be proven by~\textcolor{blue}{\cite{Xie2019SDN}}, ~\textcolor{blue}{\cite{Zhang2022VRoute}},~\textcolor{blue}{\cite{Hameed2021SDN}} and~\textcolor{blue}{\cite{Li2024underwater}}. Specifically,
Xie et al.,~\textcolor{blue}{\cite{Xie2019SDN}} proposed an SDN-enhanced blockchain TMS for CAV, where the trust calculation of vehicles was still a statistically weighted average approach. 
Hameed et al.~\textcolor{blue}{\cite{Hameed2021SDN}} demonstrated that SDN and blockchain technologies can significantly enhance the scalability of TMS.
Li et al.,~\textcolor{blue}{\cite{Li2024underwater}} implemented SDN-controlled trust evaluation for UASNs, while recent studies have extended ML-based TMS to SDN architectures in CAV.
Zhang et al., ~\textcolor{blue}{\cite{Zhang2022VRoute}} proposed 
SDN controller as an agent and the DQN approach was used to obtain optimal routing decisions, including the selection of trust nodes as part of the communication link.
  
Sub-scenario 2. \textbf{Trust evaluation of SDN mechanism.} Some works were on the trust between the SDN controller and the network application~\textcolor{blue}{\cite{aliyu2020SDNTrust}} and some works were on the trust of SDN controllers~\textcolor{blue}{\cite{Shao2023SDN}}. Besides, methods were subjective logic reasoning~\textcolor{blue}{\cite{aliyu2020SDNTrust}}, fuzzy logic models~\textcolor{blue}{\cite{Shao2023SDN}}.
However, there was a lack of discussion of specific references for CAV and ML-based TMS solutions for SDN in a survey of the trust in SDN and other 5G cutting-edge techniques~\textcolor{blue}{\cite{jorquera2023Trust_5G_Survey}}.

\subsection{TMS in Vehicle-oriented Crowdsensing}
\textbf{\textit{Scenario description:}}Crowdsensing utilizes sensors carried by a large number of participants (e.g., smartphone users, vehicles, etc.) to collect, analyze, and share data for real-time monitoring.
Entities in crowdsensing are as follows: task requestors, who release task demands;  task performers(or called workers), who use sensing entities that has terminal equipment to complete the demands and submit task data to obtain compensation; and the crowdsensing platform, through which task requestors can release task data and workers can accept task demands. In 2024, a survey of crowdsensing in smart cities was proposed ~\textcolor{blue}{\cite{Wang2024CrowdSurvey}}, where divided into three kinds, i.e., human-oriented, vehicle-oriented and infrastructure-oriented. Vehicle-oriented crowdsensing lets vehicles act as sensing entities, including opportunistic sensing vehicles(taxis and buses, which must follow a fixed trajectory), and participatory sensing vehicles(private vehicles). Our survey focuses on the trust in vehicle-oriented crowdsensing and CAV as workers. 
\subsubsection{TMS in worker selection }
\textbf{\textit{Scenario description:}}Worker selection is the crucial component for crowdsensing to control the quality of sensing services. Work selection methods mainly aim at the optimization by the effect of vehicle trajectories, urban road networks and traffic flow summarized by~\textcolor{blue}{\cite{Wang2024CrowdSurvey}}. For instance, Zhu et al.,~\textcolor{blue}{\cite{Zhu2021CrowdsensingTraj}} proposed and greedy algorithm-based vehicle recruitment based on trajectory prediction by RNN. But these methods \textcolor{orange}{did not} consider the attacks.

(1) \textit{Lessons from traditional TMS methods}.
Trust-based work selection can help filter out high-quality workers and identify and exclude potentially malicious workers against attacks. Traditional methods-based TMS for CAV were inspired by that for IoT, and have been strongly emphasizing privacy preservation from the beginning. For instance, TrustWorker~\textcolor{blue}{\cite{Gao2022TrustWorker}} was a trustworthy and privacy-preserving worker selection scheme for blockchain-based crowdsensing.
$D^{2}MTS$ in~\textcolor{blue}{\cite{Luo2024Crowd}} proposed the trust value for mobile IoT users by a weighted average approach with time weight and penalty factor. 
Li et al.,~\textcolor{blue}{\cite{Li2022Crowd}} evaluated the trust of each vehicle provider and ensured that only trusted data was for further processing and analysis, where Wasserstein GAN and differential privacy were used for the trajectory privacy-preserving scheme, but its trust assessment was still weighted averages. 
Zhao et al.,~\textcolor{blue}{\cite{Zhao2022Crowd}} chose to use an indirect reputation that is largely based on a series of historical evaluations from the CAV's past participation experiences, rather than directly requiring movement patterns, spatio-temporal distribution and sensor data analysis.
Fu et al.,~\textcolor{blue}{\cite{Fu2023BFCRI}} proposed an attributes and reputation-based worker selection scheme, where reputation was an improved multi-weight subjective logic reputation calculation model.

(2) \textit{Recent ML-based TMS methods}.
ML-based TMS has begun to be used for crowdsensing in other fields.
\begin{itemize}
    \item \textbf{GNN:} Zhang et al.,~\textcolor{blue}{\cite{Zhan2024GNNT}} proposed a GNN-based trust assessment for worker recruitment. GNN is composed of an embedding layer of Node2Vec pre-trained model, an augmented convolutional layer under expert knowledge, GCN multilayer and a fully connected layer. 
    
    \item \textbf{GAN:} In other fields, GALTrust~\textcolor{blue}{\cite{Akram2024GAL}} was a GAN and type-2 fuzzy logic-based trust classification model for spatial crowdsensing drone, which uses sequences of historical trust values as a training dataset, and also introduces a GAN-based trust redemption model to help misjudged nodes recover their trust values.
\end{itemize}

In ~\textcolor{blue}{\cite{zheng2023CITE}}, trust assessment methods covered active trust and verification trust for sensing devices as well as recommendation trust and verification trust for mobile vehicles. However, the trust value relied on counting the number of successes and failures without analyzing the content of the data packets themselves, and the weight was set manually, which limited the flexibility and accuracy of trust assessments.
With the growing popularity of blockchain for IoV, several works have proposed blockchain-based trust data dissemination. 

\subsubsection{TMS in worker incentives}
Existing incentives are categorized into different types based on various criteria. 
a) Time: ex-ante incentives and ex-post incentives~\textcolor{blue}{\cite{Liu2023BlockTMS}}.
b) Methods: Classical monetary incentive method, auction theory, game theory~\textcolor{blue} {\cite{Wang2024CrowdSurvey}}.ML-based methods should be added. 
c) Goals: privacy-preserving incentives, quality-aware incentives, reputation incentives~\textcolor{blue}{\cite{Wan2024CrowdInce}}. 
d) Serving objects: task initiators(TIs), platforms, and participants~\textcolor{blue}{\cite{Li2024CrowdIncentive}}. 
e) Blockchain: 
Thus, the worker incentives problem is defined ex-ante, focusing on ML-based methods, trust as indicators, participant-oriented and CAVs as workers.

(1) \textit{Lessons from traditional TMS methods}.
TMS methods have been integrated into traditional incentives, such as auction game~\textcolor{blue}{\cite{Xu2022Crowd}},contract~\textcolor{blue}{\cite{Dai2022Incentive}}, and Stackelberg game~\textcolor{blue}{\cite{liu2023optimalInc}}. It also became common in various scenarios, such as participants' incentives in mobile IoT crowdsensing~\textcolor{blue}{\cite{Wan2024CrowdInce}},~\textcolor{blue}{\cite{Zhang2023Crowd}}, as well as mobile IoV crowdsensing{\cite{Cai2023Incentive}}.
Specifically, a quality-aware reputation-based reward and penalty strategy was proposed to achieve dual incentives (both money and reputation) in~\textcolor{blue}{\cite{Wan2024CrowdInce}}, while a blockchain-assisted trust-preserving mechanism was introduced by a probabilistic multi-class trust assessment model in~\textcolor{blue}{\cite{Zhang2023Crowd}}. Similarly, an incentive mechanism was proposed considering security protection and data quality assurance.

(2) \textit{Recent ML-based TMS methods.}.
Additionally, a DRL-assisted incentives model has become a trend, leveraging DRL to address complex issues like multiple agents, continuous actions, and privacy protection.

\begin{itemize}
    \item \textbf{RL for worker incentives without trust.} A MADDPG-based incentive mechanism was designed for CAV crowdsensing task allocation to maximize satisfaction~\textcolor{blue}{\cite{Li2024CrowdIncentive}}.  A Q-Learning method for task allocation was proposed under spatiotemporal crowdsensing~\textcolor{blue}{\cite{Jiang2024RLCrowd}}. Zhao et al.,~\textcolor{blue}{\cite{Zhao2024DRLCroInc}} proposed a TD3-based contract design with complete information asymmetry in mobile crowdsourcing networks.A MA-DADDPG was used for solving the equilibrium policy in the multi-leader multi-follower Stackelberg game model defining the interaction between edge servers (ESs) and edge devices (EDs)~\textcolor{blue}{\cite{Zhao2023MADRLInc}}. A decentralized Soft Actor-Critic algorithm grounded in game theory (DSACG) was proposed for edge-cloud incentive mechanism~\textcolor{blue}{\cite{He2024Inc}}. A PPO-assisted VCG Auction Mechanism was proposed for long-term adaptive incentives in sustainable FL With periodical Client shifting~\textcolor{blue}{\cite{Wu2024FLInc}}.
    
    \item \textbf{RL for worker incentives with trust or other factors like trust.} Wang et al.,~\textcolor{blue}{\cite{Wang2023IncTrustCro}} proposed
    trust evaluation on ITS-Users utilizing FL and task distributors utilizing DRL, respectively. Through reasonable and fair rewards and penalties, the system encouraged all participants to raise their behavioral standards. Some studies proposed DRL-based incentives. A PPO-based method was proposed for incentivizing edge nodes participating in FL~\textcolor{blue}{\cite{Zhan2020IncFL}}. Although trust was not discussed explicitly, an incentive mechanism based on Stackelberg game theory and MA-DDPG algorithmic approach was used for data collection in platoon autonomous driving~\textcolor{blue}{\cite{Li2022DRLInc}}, where social effect and cooperative behavior were closely related to trust. It can also learn from the incentives in edge computing and FL. 
    
\end{itemize}

\subsection{TMS in CAV Platooning}
\textbf{\textit{Scenario description:}} A platoon of CAVs is defined as a string of CAVs driving in a coordinated way and maintaining extremely short inter-vehicle spacing without compromising safety, even at relatively high speeds. The development of CAV platooning aims at enhancing traffic flow efficiency, strengthening road safety, and reducing fuel consumption and emissions, thereby improving productivity for individuals and companies~\textcolor{blue}{\cite{Mar2024PlatBenefit}}. Studies on TMS in CAV platooning support critical studies, such as cooperative control algorithms, energy saving, and attack-resistant stability. We analyzed the two representative aspects in TMS. While existing literature covers traditional TMS methods, emerging approaches leveraging ML and blockchain are still underexplored.

\subsubsection{Platoon Leader Vehicle Selection}
The leader vehicle plays a pivotal role by setting the platoon's speed and trajectory, coordinating follower vehicles, and facilitating data exchange and decision-making.
Selecting a higher-trust leader ensures reliable coordination and mitigates risks from malicious or faulty vehicles.

The process is analyzed as follows, using {\cite{Li2023PlatoonCSAS} as an example. It includes four entities: User Vehicles, Platoon Leader, RSUs, and Service Provider (SP). After completing a platoon trip, user vehicles submit feedback on the platoon leader to the RSU. The platoon leader’s reputation score is calculated using a weighted average and filtering algorithm. The SP then recommends platoon leaders with high reputation scores to users to ensure the reliability of the platoon.}

Traditional TMS solutions are summarized as follows. a) Truth discovery-based TMS: It was proposed in ~\textcolor{blue}{\cite{Zhang2022Pla}} and 
~\textcolor{blue}{\cite{Cheng2023PPRT}} 
for leader vehicle selection in the CAV platoon. In ~\textcolor{blue}{\cite{Zhang2022Pla}}, pseudonym and Homomorphic encryption were proposed.In ~\textcolor{blue}{\cite{Cheng2023PPRT}}, Homomorphic encryption and data scrambling were proposed for privacy preservation.
b) Blockchain-based TMS. Ying et al.,~\textcolor{blue}{\cite{Ying2022Pla}} implemented a blockchain record of reputation values and a multi-weighted subjective logic-based TMS model. A blockchain-based security scheme was proposed in~\textcolor{blue}{\cite{Chavhan2023Platoon}} for the vehicle platoon with vehicle reputation evaluated based on its past performance, but with no specific calculation process.
c) Role-based TMS. A role-adaptive trust model for trust evaluation of platoons was proposed in~\textcolor{blue}{\cite{Chen2025Platoon}}.

\subsubsection{Trust Consensus Speed Advisory System (CSAS) in CAV Platoon}
CSAS is a special form of intelligent speed advisory system that minimizes energy consumption for a platoon of vehicles traveling in a specific area by recommending a consensus speed. TMS in CSAS ensures vehicle platoon security.
A Blockchain-Based Speed Advisory System (BSAS) {\cite{Li2022BSAS_Platoon} and Eco-Friendly CSAS {\cite{Li2023PlatoonCSAS} were proposed that use cryptography to ensure service trust and data privacy~\textcolor{blue}. Although belonging to non-trust-value TMS, the blockchain architecture, trust incentive mechanism, and privacy mechanism are helpful for trust-value-based TMS. Li et al.,~\textcolor{blue}{\cite{Li2023platoon}} designed a reputation calculation based on a multi-weighted subjective logic model for malicious node exclusion and speed recommendation.

\subsection{Advancements on the Mechanism of TMS}
In this part, the works are summarized by novelties lying in TMS mechanisms but not in supporting any services mentioned above. 
\subsubsection{ML for Node-centric TMS} Traditional TMS methods have been used, such as statistically weighted methods~\textcolor{blue}{\cite{Liu2022HDRS}},~\textcolor{blue}{\cite{Cheong2024MalNode}},~\textcolor{blue}{\cite{wang2021mobile}}. Besides, ML-based TMS are increasingly rich for CAV.

\begin{itemize}
    \item \textbf{Clustering.}Although the node trust acquisition method was based on Type-2 fuzzy logic, a K-means algorithm was proposed to take into account the distribution and pattern of trust values and to determine the trust that distinguishes between normal and malicious nodes, no longer relying on a single threshold of trust values~\textcolor{blue}{\cite{Wang2024Mali}}. A trust feature aggregation method based on k-means clustering can be first performed and then classifiers are used for trust management, such as using SVM~\textcolor{blue}{\cite{khan2023WSNTrust}}, ~\textcolor{blue}{\cite{huang2024UASN}}, and random forest~\textcolor{blue}{\cite{sagar2020KNNSVM}}.

    \item \textbf{Classifier and Regression.} Early ML-based TMS used simple neural networks such as multilayer feedforward neural networks (FNN)~\textcolor{blue}{\cite{el2020machine}},~\textcolor{blue}{\cite{Zhang2021AIT}},~\textcolor{blue}{\cite{Li2024MalIoT}} and logistic regression~\textcolor{blue}{\cite{Zhu2023HCSC}}. Then some works were on deep learning methods, such as LSTM~\textcolor{blue}{\cite{Basha2023AIDrivenMFA}}.
    Specifically, AIT~\textcolor{blue}{\cite{Zhang2021AIT}} were for both local trust equipped on vehicles and also for global trust on RSU, which  was different from non-hierarchical trust calculation in 
    ~\textcolor{blue}{\cite{el2020machine}}.
    Li et al.,~\textcolor{blue}{\cite{Li2024MalIoT}} proposed a feed-forward NN-based IoT trust prediction method, using historical trust sequences as inputs, which outperformed the traditional gray prediction model in terms of prediction accuracy, resistance to malicious attacks, and adaptability for practical applications. Some studies have been performed by defining trust metrics first and then combining the trust metrics to form a trust dataset containing the trust metric values of each node in the network and then utilizing ML algorithms~\textcolor{blue}{\cite{Du2022IUASN}}. Shafi et al.,~\textcolor{blue}{\cite{Shafi2023TrustRanking}} proposed an SVM-based trust ranking security preserving model for B5G/6G.
    TMS for CPS has been explored, but few work is on CPS for CAV. A survey on CPS security~\textcolor{blue}{\cite{harkat2024CPS}} mentions trust mechanisms but does not address CAV CPS or ML-based TMS. Azad et al.,~\textcolor{blue}{\cite{azad2024ibust}} proposed FP-Growth-based behavioral features analysis and the naive Bayesian classification method for trust decision-making in securing industrial CPS.
    
    \item \textbf{RL}. RL is used to provide feedback based on the behavior of the device and security events. 
    Trust Update Mechanism Based on Reinforcement Learning (TUMRL) was proposed in~\textcolor{blue}{\cite{He2022TrustUpdateRL}} for UASNs. Basha et al.,~\textcolor{blue}{\cite{Basha2023AIDrivenMFA}} also used RL for the trust update method for MTC communication. The higher the anomaly score, the lower the trust value. However, these two methods \textcolor{orange}{did not} specify which specific RL algorithm was used. Lin et al.,~\textcolor{blue}{\cite{lin2022drl}} proposed DDPG-based granular trust evaluation for IoT, as well as fine-grained TL-DDPG-based user trust evaluation and coarse-grained TL-DQN-based user trust evaluation.
\end{itemize}

\subsubsection{ML for Data-centric TMS}
Traditional methods have been used in TMS methods. A data-centric trust model for CAV consisted of mathematical formulas for expectation, risk, and confidence calculation in~\textcolor{blue}{\cite{Zhang2023Data}}. However, the location expectation only considers if the distance between the receiver and the sender is within a threshold, and the speed expectation only considers if the speed and predicted speed of senders are within a threshold, which cannot effectively handle Sybil attacks, data tampering attacks, and similar threats.
\begin{itemize}
    \item \textbf{Classifier.} A blockchain-based trust model was proposed ~\textcolor{blue}{\cite{wang2022dataMalV}} for vehicles deployed on RSUs, where the trust of the message content was calculated using FNN.
\end{itemize}
    
\subsubsection{ML for Hybrid TMS}
Traditional methods for hybrid TMS have been used, such as fuzzy logic~\textcolor{blue}{\cite{Hasan2023Fuzzy}}, Genetic programming (GP) ~\textcolor{blue}{\cite{aslan2023Evo}}, statistically weighted methods~\textcolor{blue}{\cite{El-Sayed2022Hybrid}}
Specifically, Ahmad et al.,~\textcolor{blue}{\cite{Ahmad2021Hybrid}} proposed that the node-centric trust is evaluated at the transport layer, and the data-centric trust is evaluated at the application layer. 

\begin{itemize}
    \item \textbf{Classifier.} Liu et al.,~\textcolor{blue}{\cite{Liu2016ART}} proposed a hybrid TM for VANET, where data trust evaluation is for traffic data, node trust includes functional trust and collaborative filtering-based recommendation trust.
    Although the trust calculation was not by ML in~\textcolor{blue}{\cite{philip2023multisource}}, a vehicle-side conflict resolution model based on LSTM was proposed serving for TMS.
\end{itemize}

\begin{table*}[h!]
\centering
\caption{Summary of ML methods in Social IoT and their vulnerabilities to trust-related attacks}
\begin{tabular}{p{0.5cm} p{0.5cm} p{1cm} p{2cm} p{2cm} p{4cm} p{5cm}}
\toprule[1.5pt]
Ref. & Time & Subject & ML Methods & Dataset & Trust-related Attacks & Drawbacks \\
\midrule
\textcolor{blue}{\cite{magdich2022res}} & 2022 & Social IoT & LR, SVM, RF, ANN, MLP, DBN & thlabsigcomm2009, Pokec & SPA, BMA, BSA, OSA, OOA, DA, WA & Not clearly described the process of trust assessment, lack of details on the simulation of trust-related attacks, simple ML methods \\
\textcolor{blue}{\cite{Marche2021MLTMSAttack}} & 2021 & Social IoT & iSVM & Santander Smart City & ME, DA, SPA, BMA, OSA, OOA, WA, Sybil attack & Lack of details on the simulation of trust-related attacks, and simple ML methods \\
\textcolor{blue}{\cite{Wang2024TrustGuard}} & 2024 & Social IoT & GNN & Bitcoin-OTC, Bitcoin-Alpha & BMA, BSA, OOA, CA & Trust-related attacks only target edges in the graph structure \\
\bottomrule[1.5pt]
\label{tab:trustatt}
\end{tabular}

\begin{minipage}{\textwidth}
\footnotesize
\raggedright
\textcolor{black}
{Notes: Self-Promoting Attack (SPA), Bad Mouthing Attack (BMA), Ballot Stuffing Attack (BSA), Opportunistic Service Attack (OSA), On-Off Attack (OOA), Discrimination Attack (DA), Whitewashing Attack (WA); Malicious With Everyone (ME); Collusion Attack(CA)};
\\
\end{minipage}
\end{table*}

\subsubsection{ML for Incentives for TMS}
Different from the discussed TMS in worker incentives for crowdsensing in ITS, a sound TMS for any service should provide incentives for encouraging nodes to provide accurate and honest recommendations or evidence of trust.
Some non-ML-based trust incentive methods have been proposed. Besides, several approaches analyzed the robustness of traditional TMS. RIETD was a reputation incentive scheme, where the fairness and reliability of RIETD were ensured through reputation updates with freshness, reputation-based revenue allocation, and defense against trust-related attacks~\textcolor{blue}{\cite{Zhao2024RepInc}}. On the other hand, although some studies discussed the robustness of the proposed non-ML-based or ML-based TMS, some studies have begun to separately discuss how to use ML methods to defend against trust-related attacks.

\begin{itemize}
    \item \textbf{ML for trust incentives.} A Q-learning-based incentive was proposed to encourage vehicles to continuously submit accurate traffic information and optimally schedule the incentive for both platform and vehicle~\textcolor{blue}{\cite{Tong2023Incentive}}, where the trust calculation was not based on ML.
    Wu et al.,~\textcolor{blue}{\cite{wu2024privacy}}
    trained trust models using FL with differential privacy and designed an incentive mechanism based on evolutionary games. Inspiration is expected from DRL-based incentives in crowdsensing and edge computing.
    \item \textbf{ML for defending trust-related attacks.} ML-based node behaviors analysis was carried out within the trust assessment process for social IoT to limit interactions with poor service providers and with attacker nodes~\textcolor{blue}{\cite{magdich2022res}}. An incremental SVM (iSVM) method with multi-parameter evaluation was proposed in a decentralized and resilient TMS for social IoT, where iSVM can defend multiple trust-related attack scenarios for dynamic knowledge learning and adapting~\textcolor{blue}{\cite{Marche2021MLTMSAttack}}. Moving from common ML methods to GNN, TrustGuard improved the accuracy and robustness of the evaluation by using similarity-based robust aggregators, robust coefficients at the spatial aggregation layer; location-aware attention mechanisms at the temporal aggregation layer; multi-head attention mechanisms; and dynamic graph segmentation~\textcolor{blue}{\cite{Wang2024TrustGuard}}. The comparison of this part is summarized in~\textcolor{blue}{Table.\ref{tab:trustatt}}.
    
    \item \textbf{ML for privacy-preserving TMS.} Some blockchain-based TMS have been proposed as privacy-preserving, such as location privacy preserving in~\textcolor{blue}{\cite{Li2021BlockTMS}},~\textcolor{blue}{\cite{Li2024BlockTMS}}, ~\textcolor{blue}{\cite{Shen2024Block}}; and identity privacy in~\textcolor{blue}{\cite{Yan2024BlockPri}},~\textcolor{blue}{\cite{Li2023TMSBlock}}. Non-blockchain-based TMS with privacy-preserving have also been proposed, focusing on trajectory privacy in~\textcolor{blue}{\cite{Li2022Crowd}}; data privacy and location privacy in~\textcolor{blue}{\cite{Liu2022Pri}}; reputation privacy and identity privacy in~\textcolor{blue}{\cite{Li2024RepPrivacy}}.%Liu2022Pri PPTM: A Privacy-Preserving Trust Management Scheme for Emergency Message Dissemination in Space–Air–Ground-Integrated Vehicular Networks；Li2024RepPrivacy RPPM: A Reputation-Based and Privacy-Preserving Platoon Management Scheme in Vehicular Networks
    However, these methods were not ML-based. ML-based privacy-preserving methods gain hot attention. Privacy-preserving graph ML~\textcolor{blue}{\cite{fu2023privacyGML}} including SMC in graph data, FL in graph data and computation at graph level,subgraph-level and node level. Cryptographic primitives like SMC, zero-knowledge proofs, HE and Functional Encryption (FE) used in privacy-preserving ML were surveyed in~\textcolor{blue}{\cite{Qian2024CP_ML}}, though these surveys rarely cover CAVs and TMSs. FL has been applied to ML-based privacy-preserving TMS, including FL-based TMS for CAV~\textcolor{blue}{\cite{wang2023Fed}}, FL with differential privacy-based TMS for cross-domain IoT~\textcolor{blue}{\cite{wu2024privacy}}. Distributed ML with differential privacy, HE and SMC are also employed to defend against privacy attacks in ML, including model inversion, membership inference and attribute inference, etc. 

\end{itemize}    

\section{Discussions}
This discussion examines three key aspects: (i) summarizing the answers to the research questions proposed in Section II, (ii) mapping the problems solved by recent studies in relation to open directions identified in prior surveys, and (iii) analyzing the barriers to implementing ML-based TMS for CAVs from theory to practice.

\subsection{Key Insights Summary in Terms of Research Questions}
The former four RQs are addressed as follows.

To address RQ1, this survey examines ML advantages and implementations in TMS for CAVs (Section IV, Parts A-D). At the data layer, ML handles heterogeneous data. At the computation layer, deep learning models complex trust relationships. At the incentive layer, algorithms optimize trust mechanisms (e.g., RL), addressing CAV-specific challenges.

To address RQ2, a comprehensive ML-based TMS evaluation framework is established (Section IV, Part E), covering traditional metrics (accuracy, efficiency) and CAV-specific needs (real-time response, privacy, interpretability, etc.).

To address RQ3, an assessment of existing work reveals gaps, particularly in dynamic topology handling and real-time performance, especially shown in Section V. It highlights key research limitations.

To address RQ4, this survey explores the integration of TMS with emerging CAV applications (e.g., trust-based platoon cooperative control and crowdsourced data collection optimization). It demonstrates improved performance and innovative service potential.

\subsection{Key Insights Summary in Terms of Solved Problems}
The majority of publications mentioned in this survey are from 2020 to 2024. Thus, key insights can be summarized that transform open directions in an ML-based trust assessment for IoT~\textcolor{blue}{\cite{wang2020survey}} in 2020 into solved problems. 

\subsubsection{Fine-grained trust evaluation based on ML}
(1) While prior work in{~\textcolor{blue}{\cite{wang2020survey}}} had limited integration of traditional and ML-based trust evaluation methods, this survey identifies three emerging approaches: i) rule/expert-based trust labeling for ML training, ii) hybrid ML-traditional prediction models using full or partial data labeling, and iii) traditional trust computation enhanced with RL and FL for dynamic updating and decision-making.

(2) While prior work in{~\textcolor{blue}{\cite{wang2020survey}}} contained limited ML-based approaches for subjective and dynamic trust, the TMS-CAV framework in this survey addresses subjective trust shown in Section III-A. At the same time, the TMS's update module enables dynamic trust by some ML-based studies, such as RL methods.

\subsubsection{Automatic feature selection and algorithm selection} 
While prior work in{~\textcolor{blue}{\cite{wang2020survey}}} highlighted the need for automated feature and algorithm selection, this survey demonstrates ML's inherent advantages in this regard, particularly through deep learning and its extensions. The ML-based studies discussed in Section V can also support this.

\subsubsection{Security and privacy}
While prior work in{~\textcolor{blue}{\cite{wang2020survey}}} emphasized the importance of security and privacy in ML-based trust evaluation, our survey systematically addresses these concerns across all TMS modules (Section V.F). Studies on CAV are few, but on static IoT and social IoT have become universal.

\subsubsection{A framework for trust evaluation as a service}  
(1) Prior work in{~\textcolor{blue}{\cite{wang2020survey}}} identified the computational and communication demands. Although early ML-based TMS studies overlooked this aspect, recent TMS-CAV studies address these constraints, as reflected in our taxonomy of TMS objectives.

(2) As capabilities initially proposed in{~\textcolor{blue}{\cite{wang2020survey}}}, recent advances in RL, transfer learning, and continual learning (Section II.E) now enable automated trust evidence collection and model optimization, though CAV implementations remain fewer than IoT applications.

\subsubsection{Different ML methods to evaluate trust} 

(1) While prior work in{~\textcolor{blue}{\cite{wang2020survey}}} noted but further explored unsupervised and semi-supervised learning for TMS, our survey provides their taxonomy (Section D). Although algorithms have improved a lot in recent years, current studies on TMS by them are at an early stage. More efforts should be made in future work.

(2) While pointing out the importance of RL for trust assessment in{~\textcolor{blue}{\cite{wang2020survey}}} without detailed analysis, our survey comprehensively examines RL-based TMS across CAV applications (Section V), though further research is needed.

\subsubsection{Integrating ML methods with other emerging technologies} 
In{~\textcolor{blue}{\cite{wang2020survey}}}, methods were expected to combine into the trust assessment, such as knowledge fragment fusion, correlation computation, fusion pattern learning, and relation discovery. \textbf{In this survey}, recent works for CAV that directly used these techniques were few. However, some principles and objectives are similar. More efforts should be made in future work.

Beyond synthesizing solved problems from prior surveys, ML-based TMS methods for IoT and CAV have demonstrated substantial technological advancements in both IoT and CAV domains since 2020. They are particularly in the following areas: decentralized trust data management, computationally efficient trust modeling, adaptive update mechanisms, incentive structure design, and attack-resilient architectures. These developments establish a strong foundation for future ML-based TMS innovations in CAV applications.

\subsection{Addressing Implementation Challenges of ML-based TMS in Real-world CAV Applications}

The current deployment of ML-based TMS in real-world CAV scenarios faces three key dilemmas and possible ML solutions are proposed.

\subsubsection{Accuracy vs. real-time performance trade-off} 

GNNs effectively model vehicle trust dynamics but face computational challenges in meeting stringent real-time requirements. Current model distillation methods enable efficient deployment at the cost of reduced accuracy, creating a pressing need for improved distillation techniques that better preserve model performance.

\subsubsection{Data silos vs. collaborative security} Although autonomous fleets can comply with GDPR through FL, cross-manufacturer data barriers hinder comprehensive global trust evaluation. Emerging solutions combining blockchain and FL may offer a breakthrough.

\subsubsection{From theory to real-world deployment} Existing TMS solutions heavily rely on simulation environments like SUMO, but real-world edge cases, such as extreme weather, expose generalization flaws in these models. Potential solutions include hybrid testing frameworks integrating SUMO simulations with physical testbeds, along with meta-learning architectures for improved generalization.

\section{Open issues and Future Directions}
The three-layer TMS-CAV framework is employed for its demonstrated capability in achieving comprehensive domain coverage, precise technology-layer alignment, and effective cross-disciplinary integration.

\subsection{Trust Data Layer}
(1) \textbf{High-quality data source.}  
Data is critical in ML, but we found that the current TMS-CAV mostly uses datasets from its own simulations, constraining the comparison of TMS-CAV using different methods. On the contrary, there are trust datasets in social networks and online social networks, but the lack of location and speed information prevents the modeling of CAV trust well. Moreover, current research on trust-related data primarily focuses on structured data such as beacons. Further exploration is urgently needed in areas like multi-source data fusion and multi-modal data fusion. This future direction has been confirmed in the latest research. Li et al.,{~\textcolor{blue}{\cite{li2025adaptive}}} proposed an adaptive multi-granularity trust management scheme for UAV visual sensor security under adversarial attacks.

(2) \textbf{Balancing computational complexity and real-time performance of TMSs.} Several strategies can be employed. 
(a) Data streamlining: Focus on collecting only the most relevant data points for CAVs but also be sufficient to guarantee the accuracy of the ML, such as vehicle behavior, network interactions, and environmental factors.
b) Data Aggregation: Use data aggregation techniques during the data collection phase to reduce the amount of data that needs to be transmitted and processed.
c) Data caching: It helps frequently accessed data to reduce the computational cost of real-time queries.
d) Data preprocessing: Cleaning, normalization, and necessary transformations are performed.
e) Feature Selection: It should reduce the input dimensions of the model by feature selection analysis.

(3) \textbf{Privacy Preservation.} 
Many distributed TMSs have used blockchain for privacy preservation. However, the transparency and traceability of blockchain may pose privacy risks, and many blockchain applications require additional privacy-preserving measures, such as the use of zero-knowledge proofs or other cryptographic techniques to hide the identities of participants, and the design of privacy-friendly address-generation and management policies. This future direction has been confirmed in the latest research~\textcolor{blue}{\cite{Liu2024BlockPrivacy}},~\textcolor{blue}{\cite{Yin2024DidTrust}}.

\subsection{Trust Calculation Layer}
(1) \textbf{Less on label.}
The majority of ML-based TMS were based on supervised learning algorithms. It relies on a large number of high-quality labeled attack data samples.  
Meanwhile, due to the continuous evolution of attackers' means and strategies, supervised learning methods such as CNN, RNN, CNN-LSTM, etc., need to constantly update the training data and re-train the models, which makes it difficult to respond to unknown attacks promptly. Efficient Semi-supervised TMS methods are needed to effectively analyze a majority of unlabeled data samples and less labeled data. 

(2) \textbf{Balancing computational complexity and real-time performance of TMS at the data level.}
Several strategies can be employed.
a) Algorithm optimization: selecting or designing algorithms with low time complexity to reduce computation time.
b) Parallel computing: Utilizing multi-core processors or distributed systems for parallel computing to increase processing speed.
c) Hardware acceleration: Use specialized hardware such as GPUs or FPGAs to accelerate computationally intensive tasks evaluated with trust.
d) Model simplification: Reduce computational complexity by using simpler models or reducing model parameters.
e) Incremental computation: Use an incremental approach for trust value updates to avoid recalculating from scratch.

To ensure CAV-TMS can be practically deployed, three key engineering constraints must be met: a) Computational Limits: The system must match onboard processor capabilities. For example, NVIDIA Drive Xavier's 10-30 TOPS performance requires GNN models under 100 MB. b) Real-Time Demands: Critical functions like emergency braking need end-to-end latency under 100ms. c) Scalability: The design must handle all traffic densities, from sparse to dense urban, without performance loss.

(3) \textbf{More efforts on CAV scenarios.}
After the survey of five scenarios, it is clear that advanced ML methods are still less used in CAVs although some studies have been widely used in social networks, static IoT, and social IoT. Besides, TMS methods are expected to be effective in more CAV scenarios, such as 
trajectory prediction,
traffic flow prediction, 
intersection signal control, non-signal control, etc. 

(4) \textbf{Detailed ML methods comparison.} 
Most studies rely on self-simulated simulations. A lack of public datasets for CAV trust causes some studies to employ advanced ML methods but lack in-depth comparisons. Future work should involve more thorough comparisons of ML-based TMS, such as evaluating RL-based approaches like DQN, DDQN, and D3QN, beyond mere parameter tuning. Additionally, comparisons with traditional TMS using the same dataset are recommended. With the growing development of security methods based on Large Language Models (LLMs), there is a need to explore LLM-based TMS, especially considering the requirements for multi-source and multi-domain data fusion.

(4) \textbf{ML-based TMS for CAV-CPS.} 
Although some studies have started exploring TMS for CPS~\textcolor{blue}{\cite{harkat2024CPS}},~\textcolor{blue}{\cite{azad2024ibust}}, CAVs are often treated as CPS. However, current TMS approaches for CAVs typically focus on the network perspective rather than addressing the broader CPS system. CPS integrates compute, network, and physical components, and TMS applied to CPS needs to consider the cross-domain requirements.

(5) Novel fields like edge intelligence, explainable ML, meta learning, and knowledge distillation.}
Edge Intelligence can assist in handling dynamic environments, multi-source data fusion, and privacy protection. Explainable ML enhances the interpretability of TMS models. Meta-learning optimizes the generalization capability of TMS. Knowledge distillation facilitates model lightweighting and efficiency improvement.

\subsection{Trust Incentive Layer}
(1) \textbf{Adaptive incentive mechanisms.}
Currently, there are several incentive studies on worker incentives in crowdsensing and edge computing, which are categorized as incentives based on traditional methods (auction, contract and game), blockchain-based incentives, DRL-based or FL-based incentives, and the combination of ML and traditional methods. However, most studies focus on general mobile scenarios, with limited attention to CAV-specific incentives, especially on trust incentives alone. There is an urgent need to develop adaptive trust incentive strategies tailored to CAVs that respond dynamically to network and environmental changes.

(2) \textbf{Trust-related attack  for ML-based TMS.}
In most current ML-based TMS studies, trust-related attacks are typically discussed in the context of the overall effectiveness of TMS. However, there is a lack of in-depth analysis on how these attacks specifically affect ML-based TMS. Given the growing importance of ML model robustness, especially against threats like data poisoning and backdoor attacks, it's crucial to examine the interplay between trust-related attacks, adversarial attacks on ML, and cyber attacks. For instance, collusion attacks, a common type of trust-related attack, are often triggered by Sybil attacks, and they can have a similar impact on ML systems as data poisoning.

\section{CONCLUSIONS}
Trust mechanisms are essential for secure data exchange, reliable decision-making, and defense against attacks, evolving from simple modules into comprehensive systems known as trust management systems. Traditional rule-based TMS has been significantly enhanced by ML, which brings advantages like handling large-scale, high-dimensional data and adaptive learning capabilities.

An innovative three-layer ML-based TMS framework designed for CAVs is introduced in our survey, integrating the trust data layer, trust computation layer, and trust incentive layer. The strengths of ML are leveraged in each layer, offering a holistic solution for TMS in CAVs. A six-dimensional objective framework is also proposed, aligning with the unique characteristics and requirements of CAVs, serving as a benchmark for evaluating and comparing various ML approaches. Additionally, recent advancements shown by studies in TMS across five traffic scenarios and recent advancements on TMS mechanisms are surveyed.

As an exploratory survey, this survey acknowledges that ML techniques and CAV technologies are rapidly evolving. While our proposed framework is designed to be scalable and adaptive, ML-based TMS still faces critical challenges in rigorous CAV environments, including the robustness to dynamic conditions, the real-world validation gap, explainability and safety, etc. Despite these challenges, this survey provides a foundational roadmap for future research. As CAV and ML technologies mature, addressing these limitations will be pivotal to achieving trustworthy and resilient CAVs.

\ifCLASSOPTIONcaptionsoff
  \newpage
\fi




\bibliographystyle{IEEEtran}
\bibliography{IEEEabrv,Bibliography}

\begin{IEEEbiography}[{\includegraphics[width=1in,height=1.25in,clip,keepaspectratio]{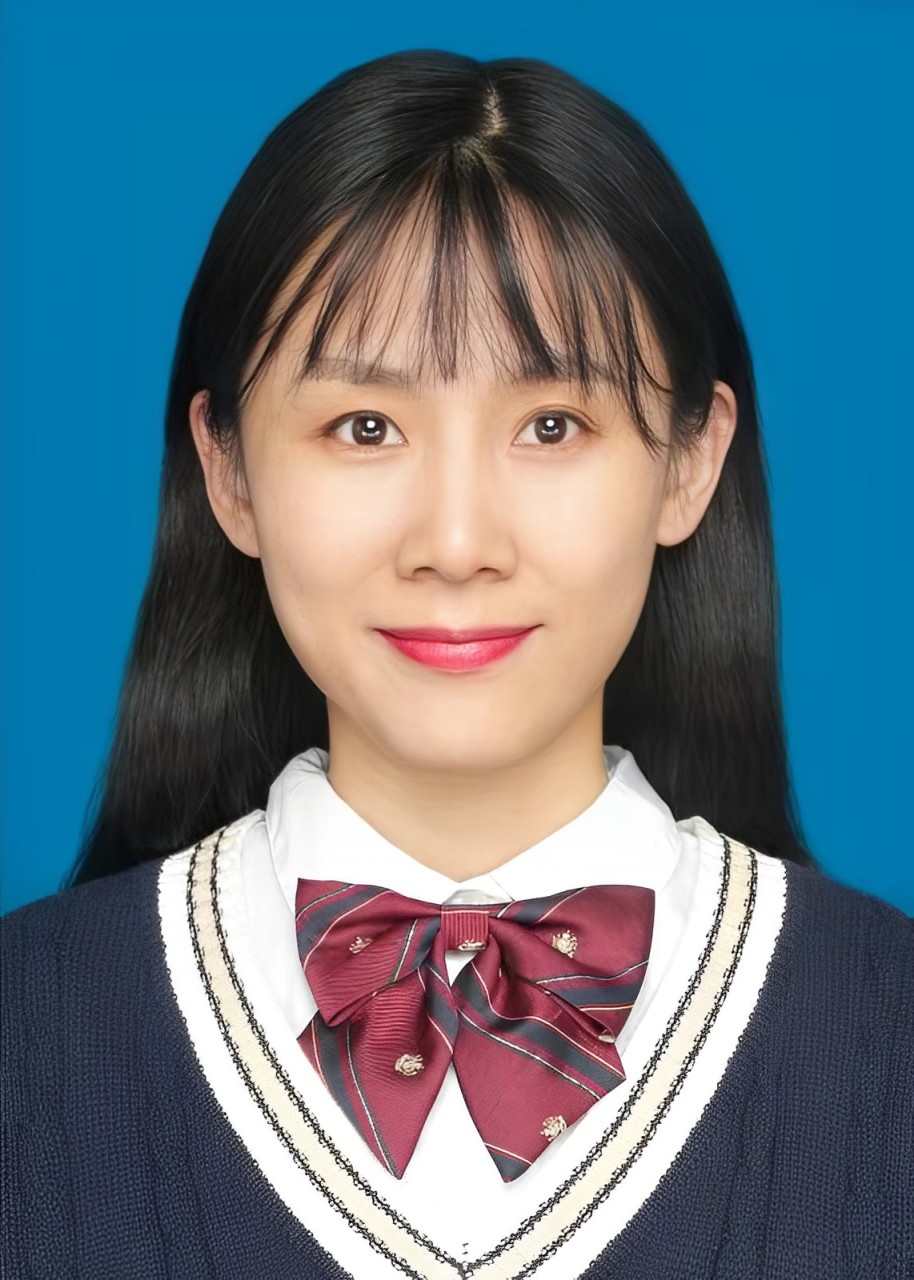}}]{Qian Xu}
received bachelor’s and Master’s degrees in Traffic Information Engineering and Control from Lanzhou Jiaotong University in 2018 and 2021, respectively. She received the Ph.D. degree in Transportation Engineering in 2025 at Tongji University, Shanghai, China. the Ph.D. degree in Transportation Engineering in 2025 at Tongji University, Shanghai, China. Her research focuses on security and privacy techniques for connected and automated vehicles, especially machine learning-based trust management systems.
\end{IEEEbiography}

\begin{IEEEbiography}[{\includegraphics[width=1in,height=1.25in,clip,keepaspectratio]{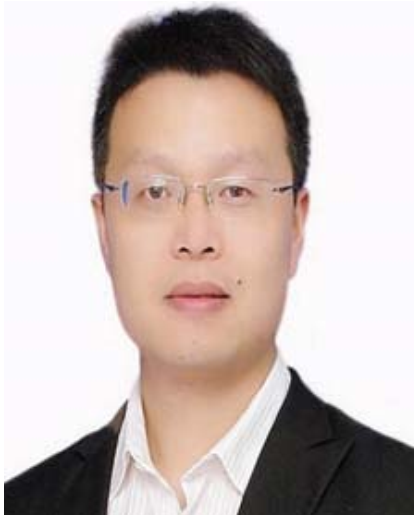}}]{Lei Zhang} received the Ph.D. degree from the Chinese Academy of Sciences in 2008, and the postdoctoral program from Tsinghua University in 2010. He is currently a professor and Ph.D supervisor at Tongji University, Shanghai. He is one of the editorial board members of High-Confidence Computing. His current research interests include intelligent information processing and applications for spatio-temporal data analysis. 

\end{IEEEbiography}

\begin{IEEEbiography}[{\includegraphics[width=1in,height=1.25in,clip,keepaspectratio]{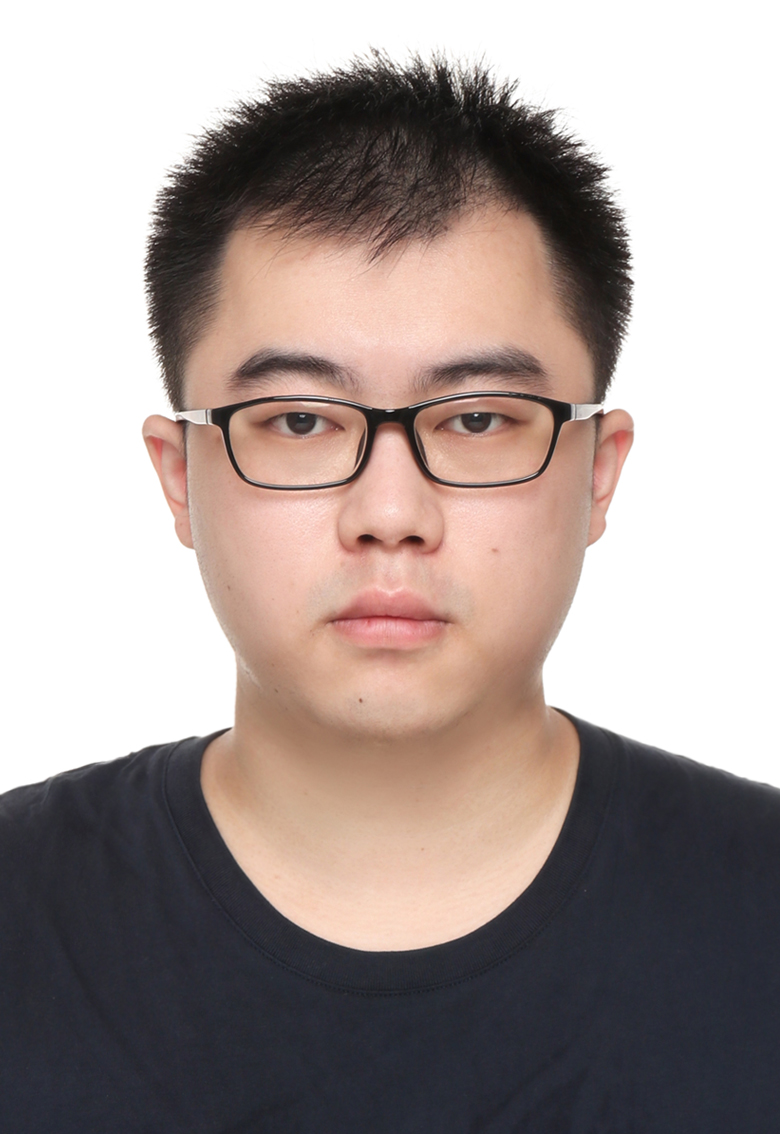}}]{Yixiao Liu} received a Bachelor’s degree in Mechanical Engineering from Tongji University in 2021 and a Master’s degree in Data Science from City University of Macau in 2023, and is currently pursuing a Ph.D in Intelligent Science and Technology from Tongji University since 2023. His research interests include spatio-temporal data analysis, trustworthy connected and automated vehicles, and multi-agent
systems.
\end{IEEEbiography}

\vfill


\end{document}